\documentclass{article}

\usepackage{PRIMEarxiv}

\usepackage[utf8]{inputenc} 
\usepackage[T1]{fontenc}    
\usepackage{hyperref}       
\usepackage{url}            
\usepackage{booktabs}       
\usepackage{amsfonts}       
\usepackage{nicefrac}       
\usepackage{microtype}      
\usepackage{lipsum}
\usepackage{fancyhdr}       
\usepackage{graphicx}       
\usepackage{amsmath}
\usepackage{subcaption}
\usepackage{wrapfig}
\usepackage{natbib}
\usepackage{comment}
\usepackage{rotating}
\usepackage[table]{xcolor}
\usepackage{soul}
\sethlcolor{green}
\graphicspath{{resources/}}     

\title{Causality-enhanced Decision-Making for Autonomous Mobile Robots in Dynamic Environments}

\author{
 Luca Castri \\
  University of Lincoln\\
  \texttt{lcastri@lincoln.ac.uk} \\
   \And
 Gloria Beraldo \\
  National Research Council of Italy\\
  University of Padua\\
  \texttt{gloria.beraldo@cnr.it} \\
  \And
 Nicola Bellotto \\
  University of Padua\\
  University of Lincoln\\
  \texttt{nbellotto@dei.unipd.it} \\
}

\begin{document}
\maketitle

\begin{abstract}
The growing integration of robots in shared environments---such as warehouses, shopping centres, and hospitals---demands a deep understanding of the underlying dynamics and human behaviours, including how, when, and where individuals engage in various activities and interactions. This knowledge goes beyond simple correlation studies and requires a more comprehensive causal analysis. By leveraging causal inference to model cause-and-effect relationships, we can better anticipate critical environmental factors and enable autonomous robots to plan and execute tasks more effectively.
To this end, we propose a novel causality-based decision-making framework that reasons over a learned causal model to assist the robot in deciding when and how to complete a given task. In the examined use case---i.e., a warehouse shared with people---we exploit the causal model to estimate battery usage and human obstructions as factors influencing the robot’s task execution. This reasoning framework supports the robot in making informed decisions about task timing and strategy.
To achieve this, we developed also PeopleFlow, a new Gazebo-based simulator designed to model context-sensitive human-robot spatial interactions in shared workspaces. PeopleFlow features realistic human and robot trajectories influenced by contextual factors such as time, environment layout, and robot state, and can simulate a large number of agents. 
While the simulator is general-purpose, in this paper we focus on a warehouse-like environment as a case study, where we conduct an extensive evaluation benchmarking our causal approach against a non-causal baseline. Our findings demonstrate the efficacy of the proposed solutions, highlighting how causal reasoning enables autonomous robots to operate more efficiently and safely in dynamic environments shared with humans.
\end{abstract}


\section{Introduction}

Autonomous mobile robots are a game changer for the progress and development of various sectors, including industry, logistics, agriculture, and healthcare.
However, operating in dynamic environments and workspaces shared with people introduces several challenges. In particular, unexpected events and human encounters can negatively affect the capacity of a robot to fulfil its task safely and autonomously.
Understanding the causal effect of the robot's actions enables it to better assess and choose them, which is crucial for improving both the efficiency and safety of its operations. This motivates the need for intelligent solutions to reason about the causal relationships between the environment and human-robot interactions, improving robot's efficiency and guaranteeing human safety. Unsurprisingly, the importance of cause-and-effect reasoning in AI and robotics has become strategically important for national and international research programmes\footnote{euRobotics, ``A Unified Vision for
European Robotics'', \url{https://eu-robotics.net/strategy/}}\textsuperscript{,}\footnote{National Robotics Initiative 3.0, ``Innovations in Integration of Robotics'', \url{https://www.nsf.gov/pubs/2021/nsf21559/nsf21559.htm}}\citep{euRobotics2024UnifiedVision,nsf21-559}.

Knowing the causal structure of a system provides a significant advantage in many robotics applications. Indeed, research in causal inference, which includes causal \textit{discovery} and \textit{reasoning}, has gained increasing attention in recent robotics literature~\citep{hellstrom2021relevance,brawer_causal_2021,cao_reasoning_2021,castri2022causal,castri2023enhancing,Katz2018,Angelov2019,Lee2022,cannizzaro2023towards,cannizzaro2023towardsdrones,cannizzaro2023car,love2024would,love2024towards}.
However, despite this growing attention, a notable gap persists between the potential benefits of modern causal inference and its actual application in real-world robotics. 
In fact, to the best of our knowledge, no prior work has explored causal inference for robot decision-making in the context of dynamic environments and human-aware navigation. The latter in particular typically relies on predictive models of human motion, which are often reactive and do not take into account robot and human goals, or how contextual factors can influence them. As a result, autonomous robot navigation in these environments can lead to potential safety and efficiency issues.

\begin{figure}[!t]
\centering
\includegraphics[width=.75\columnwidth]{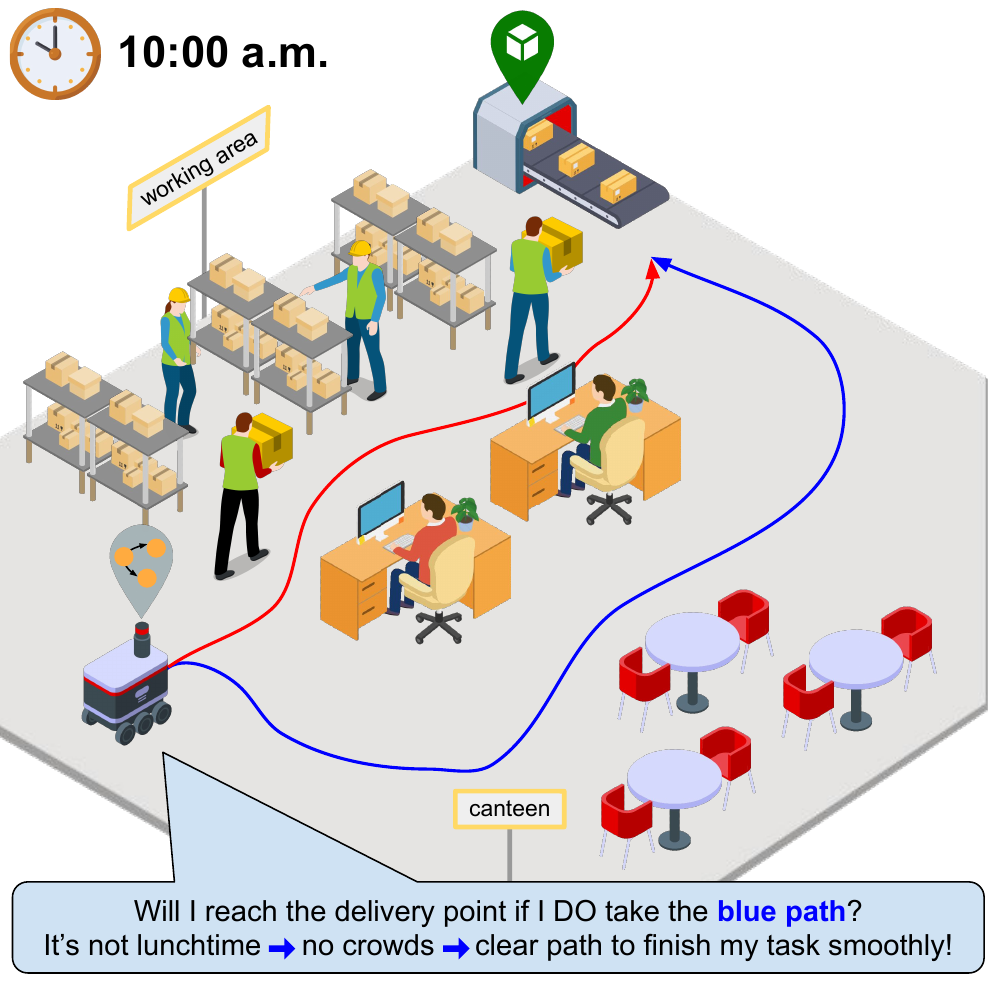}
\caption{A mobile robot reasoning on the causal model of human spatial behaviours in a warehouse environment to navigate safely and efficiently.}
\label{fig:intro}
\end{figure}
Consider for example a mobile robot in a warehouse shared with humans, as depicted in Fig.~\ref{fig:intro}. The robot must reach a target location while accounting for how humans behave within the shared space, considering the layout of the environment and other contextual factors, such as the time of day. Discovering the cause-and-effect relationships among these factors enables the robot to reason about them, making it aware of the mechanisms driving human behaviour and allowing it to make decisions based on causal knowledge. As shown in Fig.~\ref{fig:intro}, thanks to its causal model the robot infers that, although the path through the canteen is longer than the direct route through the working area, it is the safest and most efficient option \textit{because} it is not lunchtime and the canteen is less crowded. By choosing this path, the robot can efficiently reach its target position, minimising the number of unexpected obstacles and the risk of collisions with humans.

This paper presents a causality-enhanced decision-making framework to model relevant robot and human states within a shared workspace, enabling causal reasoning that improves task-execution efficiency and safety. We demonstrate that with our framework a mobile robot can reconstruct the underlying causal relationships between autonomous navigation, human spatial behaviour, and surrounding environment from data. By leveraging this knowledge, it can efficiently complete its tasks while ensuring that human-robot spatial interactions~(HRSIs) remain as safe as possible.
In summary, our contributions are as follows:
\begin{itemize}
\item a novel, end-to-end framework for robotic decision-making that, for the first time, integrates causal discovery from sensor data with causal reasoning for motion planning within the Robot Operating System~(ROS) to improve task-execution efficiency and safety in dynamic human-shared environments;

\item a new Gazebo-based simulator, PeopleFlow\footnote{PeopleFlow GitHub repository: \url{https://github.com/lcastri/PeopleFlow}\label{foot:peopleflow}}, which generates complex spatial behaviours where agents adapt both reactively to dynamic changes and proactively by adjusting their goals based on contextual factors (e.g., time of day and scheduled tasks), as shown in Fig.~\ref{fig:intro};

\item a thorough experimental evaluation of the causal framework in a complex warehouse simulation to demonstrate its feasibility and potential for real-world robotic applications in comparison to a state-of-the-art non-causal baseline.
\end{itemize}
PeopleFlow is available as a Docker image on GitHub:
~\url{https://github.com/lcastri/PeopleFlow}.

The paper is structured as follows: 
basic concepts about causal discovery and related work in the robotics fields are presented in Section~\ref{sec:related}; 
Sections~\ref{sec:causal-inference}~and~\ref{sec:simulator} explains the details of our causality-based decision making framework and PeopleFlow; 
Section~\ref{sec:exp} presents the application and results of our approach in a simulated warehouse-like environment; 
finally, Section~\ref{sec:conclusion}
concludes the paper discussing achievements and future work.

\section{Related Work}\label{sec:related}
The synergy between causality and robotics offers mutual benefits. Causality leverages robots' physical capabilities for interventions, while robots utilise causal models to gain a deeper understanding of their environment. This synergy has led to increased attention towards causal inference in various robotics applications. Causal inference comprises two main phases: causal discovery and causal reasoning.

\paragraph{\textbf{Causal Discovery in Robotics}}
Causal discovery involves identifying the underlying causal structure from data, either through observational methods, interventional experiments, or a combination of both. It aims to uncover cause-and-effect relationships between variables rather than mere correlations. 
Numerous methods have been developed for causal discovery from both static and time-series data~\citep{glymour_review_2019,                     assaad2022survey}. However, for robotics applications, where time-series sensor data is common, methods tailored specifically for time-dependent causal discovery are essential. Among the various approaches available in the literature, the time-series variation of the Peter and Clark (PC)~\citep{spirtes2000causation} algorithm, such as PC Momentary Conditional Independence (PCMCI)~\citep{runge_causal_2018}, has been widely used in many applications, including robotics~\citep{runge_detecting_2019,saetia_constructing_2021,castri2022causal}. 
PCMCI has recently been extended with several enhancements. PCMCI\textsuperscript{+}~\citep{runge2020discovering} enables the discovery of simultaneous dependencies. Filtered-PCMCI~(F-PCMCI)~\citep{castri2023enhancing} integrates a transfer entropy-based feature-selection module to improve causal discovery by focusing on relevant variables. Joint-PCMCI\textsuperscript{+}~(J-PCMCI\textsuperscript{+})~\citep{pmlr-v216-gunther23a} supports causal structure learning from multiple observational datasets using context variables. Latent-PCMCI~(LPCMCI)~\citep{gerhardus2020high} allows causal discovery in the presence of latent confounders. CAnDOIT~\citep{castri2024candoit}, which stands for Causal Discovery with Observational and Interventional Data from Time Series, extends LPCMCI by integrating both observational and interventional data for improved causal discovery. 
Finally, PCMCI and F-PCMCI have been integrated into ROS-Causal~\citep{castri2024roman}, a ROS-based framework that provides causal discovery only, facilitating their use in robotic systems but without supporting causal inference or reasoning.

\paragraph{\textbf{Causal Reasoning in Robotics}}
Once the causal structure is established, causal reasoning enables inference, prediction, and decision-making by leveraging the discovered causal relationships to estimate the values of variables within the causal model, assess the effects of interventions, and evaluate counterfactual scenarios.
Structural Causal Models~(SCMs) have been employed to understand how humanoid robots interact with tools~\citep{brawer_causal_2021}. PCMCI and F-PCMCI have been used to establish causal models for underwater robots navigating toward target positions~\citep{cao_reasoning_2021} and to model and predict human spatial interactions in social robotics~\citep{castri2022causal,castri2023enhancing}. The latter represent the first approaches to applying causal discovery for modelling HRSIs and reasoning in a forecasting task. However, they did not propose an end-to-end causal framework, nor were the predictions exploited for decision-making in human-aware robot navigation, as we do in this paper.
Moreover, causality-based approaches have been explored in various robotics domains, including robot imitation learning, manipulation and explainable HRI~\citep{Katz2018,Angelov2019,Lee2022,love2024would,love2024towards}.

Despite the growing use of causality in robotics, most existing approaches primarily focus on causal discovery and the prediction of time-series variables within the causal model. While some recent works~\citep{cannizzaro2023towards,cannizzaro2023towardsdrones,cannizzaro2023car} have explored causal reasoning for robot manipulation tasks and drone applications, the use of causal models for decision-making in human-centred environments remains largely unexplored.
In contrast, our work focuses on integrating causal reasoning into the robot’s decision-making process in dynamic environments shared with humans. By leveraging causal inference, our framework enables robots to make causality-aware decisions, enhancing both task execution efficiency and human safety in HRSI scenarios.

\paragraph{\textbf{Human Robot Spatial Interaction}}
Several studies have highlighted the importance of considering the coexistence of humans and robots in shared environment, whether they are collaborating or performing individual tasks~\citep{mukherjee2022survey,vasconez2019human,jahanmahin2022human}. In this work, we focus on the spatial aspect of the interactions between human-human, human-robot and human-object~\citep{dahiya2023survey}. Recently, various methods have been proposed to model the relations between human motion behaviours and spatial interactions~\citep{dondrup2015computational,mghames2023neuro,mghames2023qualitative}. However, these approaches have not explicitly considered causal relationships between spatial variables. Some recent studies~\citep{Liu2022,castri2022causal,castri2023enhancing} have developed causal models for modelling HRSI.

Despite these advances, none of the existing works explicitly account for contextual factors—such as the influence of surrounding agents, dynamic environmental conditions, and task-specific constraints—when modelling HRSIs. These factors can alter the relationships between variables, causing interactions to strengthen, weaken, or even emerge or disappear depending on the specific context in which the system is observed. Analysing these dependencies requires diverse datasets covering various environments, agent behaviors, and tasks, but such data are currently unavailable. Collecting them across all possible conditions would be impractical and highly time-consuming.
%

Most existing datasets primarily focus on human motion trajectories rather than capturing how contextual elements shape human-robot interactions. For instance, TH{\"O}R~\citep{thorDataset2019} and its extension TH{\"O}R-MAGNI~\citep{schreiter2024thor} provide motion trajectories recorded in controlled indoor environments using ceiling-mounted cameras, while the ATC Pedestrian Tracking dataset~\citep{brvsvcic2013person} tracks pedestrians in a shopping mall atrium using 3D range sensors. Similarly, the JackRabbot Dataset and Benchmark (JRDB)~\citep{martin2021jrdb} captures human poses from a mobile robot's ego-perspective using RGB cameras and 3D LiDAR.
While these datasets offer valuable insights into human motion patterns, they do not include contextual variability, such as how different environmental settings, external agents, or task constraints influence interaction dynamics.

To address this limitation, simulation environments offer a practical solution for generating diverse, controlled, and repeatable HRSI scenarios. By systematically varying contextual factors, simulators offer the opportunity to explore how different conditions may influence human-robot interactions, helping to address some of the limitations of existing datasets.
SEAN 2.0~\citep{tsoi2022sean2} provides a high-fidelity, Unity-based simulation environment tailored for pedestrian interactions, making it well-suited for training vision-based robotics algorithms in dynamic social settings. In contrast, MengeROS~\citep{aroor2017mengeros}, CrowdNav~\citep{chen2019crowd}, and SocialGym~\citep{holtz2022socialgym} focus on 2D grid-based visualisations, which, while useful for algorithmic benchmarking, may lack the realism needed to model complex human behaviours in rich environments. ROS-Causal\_HRISim\footnote{\url{https://github.com/lcastri/ROS-Causal_HRISim}\label{foot:roscausal_hrisim}}~\citep{castri2024roman} takes a different approach by enabling causal analysis in HRI scenarios, allowing both observational data collection and targeted interventions involving humans and robots.

%
Our new Gazebo-based simulator, PeopleFlow, extends our previous work ROS-Causal\_HRISim to support context-sensitive human-robot spatial interactions. Unlike existing simulators, it does not only record HRSI data across diverse contexts but also models and dynamically assigns the roles of humans and robots based on probabilistic destination/goal station selections conditioned on context (e.g., time of the day, tasks to accomplish). This enhanced simulation framework offers the ability to record HRSI data across different contextual scenarios, in which people and robot have a role that depends on the context, enabling a more comprehensive analysis of how environmental conditions, agent behaviours, and task constraints influence human-robot interactions.

\section{Causality-based Robot Decision Making Framework}\label{sec:causal-inference}
\begin{figure}[t]\centering
\includegraphics[trim={0cm 0cm 0cm 0cm}, clip, width=\textwidth]{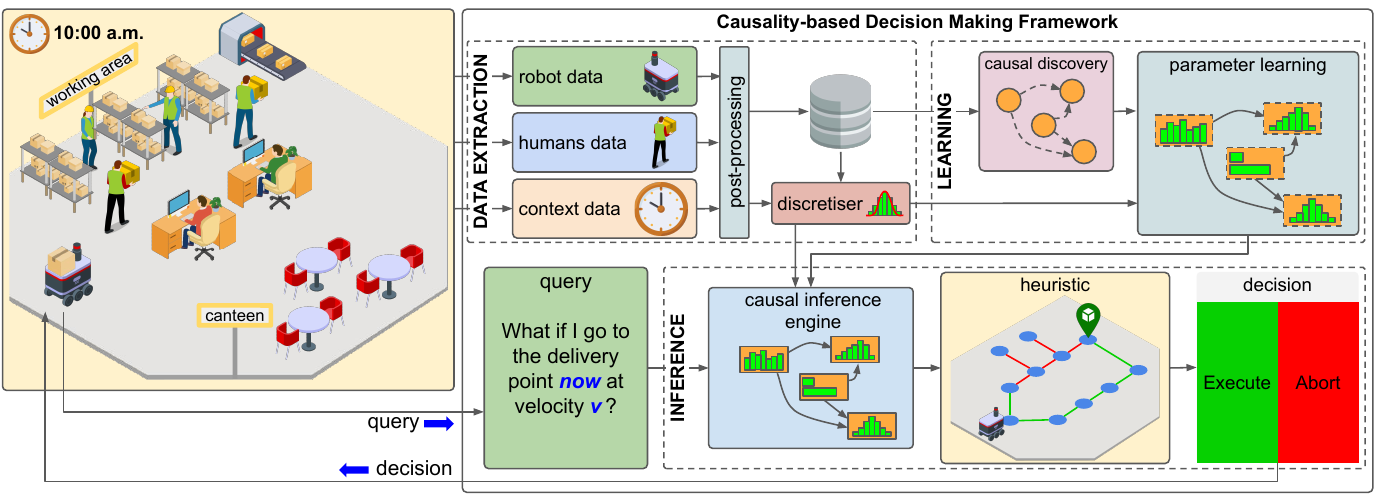}
\caption{Block scheme of the causality-based decision making framework, consisting of three main pipelines:
\emph{(i)~Data Extraction} which gathers and preprocesses data from the observed scenario;
\emph{(ii)~Learning} that retrieves the causal model describing the observed scenario and exploits its structure along with extracted data to learn the parameters of the causal inference engine;
\emph{(iii)~Inference}, which uses the learnt causal inference engine to estimate relevant quantities for determining a path to complete a task and deciding in advance whether execute it or abort it.}
\label{fig:causal-inference}
\end{figure}
Our approach introduces a novel, end-to-end framework for robotic decision-making, enabling an autonomous robot to reason about cause and effect for safer and more efficient navigation in dynamic environments. The core of this framework is a {\em causal inference engine}~\citep{pearl2018book} that allows a robot to determine a path for completing a given task (e.g., delivering an object to a designated location), looking for a trade-off among relevant factors in its operating environment. These factors may include the current state of the environment (e.g. people crowding an area at a specific time) or a robot-related constraint, such as battery level, that could influence task execution. Additionally, the framework should be capable of preemptively deciding whether to initiate or abort a task to avoid failures or inefficiencies.

While prior work has used causal reasoning for time-series prediction~\citep{castri2022causal,castri2023enhancing}, manipulation tasks~\citep{cannizzaro2023towards} and autonomous drone application~\citep{cannizzaro2023towardsdrones}, our framework is the first to integrate these concepts into an end-to-end system for human-aware navigation. The primary contribution of this paper is this novel end-to-end causal architecture, which not only employs state-of-the-art algorithms to translate raw sensor data into a causally-based plan but is also architecturally designed to support long-term autonomy. This distinction is crucial: while a ``two-stage'' pipeline (offline causal discovery followed by a separate inference engine) might achieve similar initial performance, it is inherently static and requires human intervention for any single change. Our unified framework (shown in Fig.~\ref{fig:causal-inference}), being designed as a modular system, provides the foundation to overcome this dependency on human intervention, which is the key to enhancing long-term autonomy. For instance, if the environment evolves (e.g., new worker schedules or spatial layouts), a two-stage pipeline with a ``frozen'' causal inference engine leads to degraded performance until a human manually re-runs the entire discovery and deployment process. The proposed architecture could be easily extended to eventually manage its own model lifecycle by, for example, autonomously detecting when the engine is outdated and re-running the learning pipeline to adapt to the evolving environment.

Deciding which path to take to reach a destination allows the robot to operate more efficiently while ensuring human-robot interactions remain as safe as possible. For instance, consider a scenario, shown in Fig.~\ref{fig:intro}, where the robot is in a warehouse environment shared with people and needs to travel from its current position to the delivery point, with two possible paths. The first route (red path in Fig.~\ref{fig:intro}) is the shortest but also the busiest, meaning the robot may need to spend significant time and energy navigating around people, increasing both the risk of collisions and battery consumption. In contrast, the second route is longer but completely clear (blue path in Fig.~\ref{fig:intro}). Despite being physically longer, this second path could enable the robot to reach its destination more safely, more quickly and with lower battery consumption since it avoids people.

Building our framework involves multiple steps, which can be broadly categorised into three main blocks: \emph{data extraction}, \emph{learning}, and \emph{inference}, as illustrated in Fig.~\ref{fig:causal-inference}.

\subsection{Data Extraction Pipeline}\label{sec:causal-inference-data}
The \emph{data extraction} phase is the first block of our framework, providing input to the learning and inference components. This phase is responsible for the continuous extraction of data from the observed scenario, capturing information related to the robot, humans, and contextual factors.  

For the learning pipeline, the data must be preprocessed to facilitate the discovery of the causal model and the construction of the causal inference engine. The first step in this process is subsampling. Given the large volume of collected data, subsampling is applied to reduce the dataset size while preserving essential information. This step ensures that the dataset remains computationally manageable without compromising the integrity of the data distribution.
Since our dataset consists of multiple time-series, we determine the appropriate subsampling rate based on the Nyquist-Shannon sampling theorem to prevent information loss. Specifically, we first analyse the spectral content of each time-series to identify its highest frequency component (bandwidth $BW_{\text{max}}$) and then set the subsampling rate as ${f_s \geq 2 \cdot BW_{\text{max}}}$ where $f_s$ is the subsampling frequency and $BW_{\text{max}}$ is the highest significant frequency present in the dataset. This ensures that the subsampled data retains all relevant dynamics.

Next, the data undergoes post-processing to extract relevant time-series information from the raw collected data, which is then stored for causal discovery analysis.  
To construct the causal inference engine, the extracted data must be compatible with inference mechanisms that often rely on discrete data representations. Therefore, we apply a discretisation step using pyAgrum\footnote{\url{https://pyagrum.readthedocs.io/en/1.17.2/}\label{foot:pyagrum}}’s built-in discretiser to transform continuous variables into discrete states while preserving essential statistical properties. Specifically, we use the {\em quantile}~\citep{dimitrova2010discretization} and {\em elbow method}~\citep{dangeti2017statistics} approaches for discretisation. The quantile approach is particularly suitable when data is skewed or non-uniformly distributed, as in our case, whereas the elbow method helps to automatically determine the number of bins per variable.

Once the dataset has been processed, it is ready to be fed into the learning and the inference pipelines (see Fig.~\ref{fig:causal-inference}).  
For the latter, the data is continuously captured and post-processed as described above. However, in this case, the data is not stored but instead discretised and directly used for real-time inference.  
In the following sections, we provide a more detailed explanation of the learning and inference pipelines. 

\subsection{Learning Pipeline}\label{sec:causal-inference-learning}
The \emph{learning} phase comprises both causal discovery and parameters learning. First, it first uncovers the underlying causal relationships governing human and robot behaviours in the working environment. 
This process assumes a predefined set of relevant variables based on domain knowledge.
Once the causal structure is identified, the model's parameters are estimated to enable reasoning over the retrieved causal relationships. 

As previously explained, the causal discovery block takes as input the post-processed information from the data pipeline and performs the causal analysis to uncover the underlying causal structure governing the observed scenario. The analysis is conducted using CausalFlow\footnote{\url{https://github.com/lcastri/causalflow}}, a Python library that provides a collection of methods for causal discovery from time-series, as reviewed in Sec.~\ref{sec:related}. Embedding CausalFlow into our framework makes it adaptable to various scenarios. The choice of the causal discovery method depends on the dataset characteristics. For basic causal discovery, approaches such as PCMCI, DYNOTEARS, and VARLiNGAM can be applied. When the dataset requires feature selection, F-PCMCI is used, whereas L-PCMCI is more suitable when latent variables are present. If the dataset contains a mix of observational and interventional data, CAnDOIT is the appropriate method. In our case, since the dataset contains contextual variables, we use J-PCMCI\textsuperscript{+}~\citep{pmlr-v216-gunther23a}. 
Given the data structure in our work---time-series with system variables and independent context variables---J-PCMCI\textsuperscript{+} is uniquely suited for this analysis. It is specifically designed to correctly model context variables as external factors that affect the system without being influenced by it. Applying other causal discovery methods from observational data, such as DYNOTEARS~\citep{pamfil2020dynotears}, VARLiNGAM~\citep{hyvarinen2010estimation}, tsFCI~\citep{entner2010causal}, PCMCI~\citep{runge_causal_2018}, F-PCMCI~\citep{castri2023enhancing}, LPCMCI~\citep{gerhardus2020high}, would incorrectly treat these external factors as standard system variables, thereby misrepresenting the underlying scenario. For this reason, we use only J-PCMCI\textsuperscript{+} in the following evaluation.

Inspired by~\citet{raichev2024estimating}, the causal structure derived in the previous block is used together with the discretised data to learn the parameters of the causal inference engine. For this, we again employ pyAgrum\footref{foot:pyagrum} for estimating the conditional probability distributions associated with each node in the causal graph, ensuring consistency with the causal structure identified in the discovery step. Unlike the approach in~\citep{raichev2024estimating}, which employs the Expectation-Maximization (EM) algorithm, we use Maximum Likelihood Estimation (MLE)~\citep{myung2003tutorial}, as our dataset does not contain missing data.
These steps form the \emph{learning pipeline} and are shown in Fig.~\ref{fig:causal-inference}.

\subsection{Inference Pipeline}\label{sec:causal-inference-inference}
Once built, the causal inference engine can be used by the robot to make decisions based on inferred information.
It is important to note that the causal discovery and learning parameters steps are performed only once and are not part of the real-time inference pipeline. Indeed, in this step the robot uses the already-learned causal inference engine to make causality-aware decisions based on new observations, leveraging the relationships encoded in the causal model.
Specifically, the \emph{inference pipeline} begins with a query from the robot at the start of an assigned task. Before proceeding, the robot must first gather information that is crucial for successfully accomplishing the task. As previously mentioned, this information can be related to the environment in which the robot operates or to its internal state, both of which influence task completion. 

The robot's query is formulated as an intervention, representing an action it considers taking to accomplish the task (e.g., {\em What if I do(X)?}). Through the causal inference engine, which has been previously learnt, the inference pipeline evaluates the potential consequences of this action by estimating how relevant factors would change as a result of an intervention or condition. The output of this process is the conditional probability distribution of the queried factor, taking into account causal links within the causal model. Once again, this inference step is performed using pyAgrum\footref{foot:pyagrum}, which provides a full implementation of do-calculus~\citep{pearl2009causality} for this step~\citep{ducamp2020agrum}. The estimated value of the factor is then obtained by computing its expectation over the inferred probability distribution.
The estimated value is finally embedded in a heuristic used by the A\textsuperscript{*} algorithm to define the robot's path. If a valid path is found under certain conditions, the inference pipeline returns a decision to proceed with the task using the selected path. In contrast, if no valid path is found, an abort signal is returned to the robot.
These steps constitute the \emph{inference pipeline} and are shown in Fig.~\ref{fig:causal-inference}. A use-case implementation of the framework is presented in the following section, including an example of causal inference for decision making in Sec.~\ref{sec:simulator-causal}.

\section{Simulating Contextual Human-Robot Spatial Interactions}\label{sec:simulator}
To address the limitations of existing datasets and simulators, which overlook contextual dependencies in HRSI, we introduce \textit{PeopleFlow}, a simulation framework designed to capture and model context-sensitive behaviours in human-robot and human-human interactions.

PeopleFlow enables the creation of realistic, diverse, and repeatable spatial interaction scenarios by incorporating contextual factors that dynamically influence human and robot behaviours. These contextual factors---such as time, location, and environmental obstacles---are crucial in shaping the interactions dynamic.
In addition, as highlighted in previous work~\citep{castri2024roman,castri2024ros}, such simulation environments not only support data generation but also provide a controlled setting for verifying the correctness of causal models and evaluating the experimental procedures used to extract them. This helps reduce both time wasted on ineffective experiments and unnecessary financial costs.

\subsection{PeopleFlow}\label{sec:simulator-env}
\begin{figure}[t]
    \centering
    \includegraphics[trim={0.0cm 0cm 0.0cm 0cm}, clip, width=\textwidth]{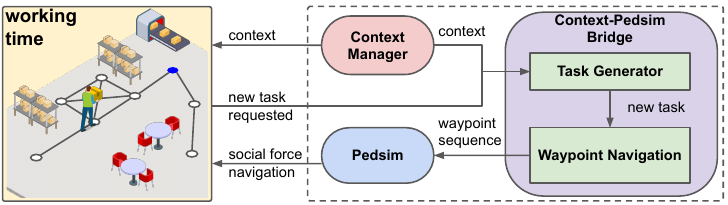}
    \caption{Overview of PeopleFlow’s context-aware agent behaviour strategy. The Context Manager node handles the contextual factors—in this example, time. After the agent picks up a box from a shelf, it requests a new task. The Context-Pedsim Bridge node receives the task request and generates a new task for the agent depending on the current context. Since it is working-time, the agent is tasked to bring the box to the delivery point. A corresponding sequence of waypoints is generated and passed to Pedsim, which navigates the agent to the goal using the social force model. The agent’s goal is marked with a blue circle, while white circles represent the remaining waypoints.}
    \label{fig:simulator-approach}
\end{figure}
We developed PeopleFlow by building upon our previous work\footref{foot:roscausal_hrisim}\citep{castri2024roman}. The latter is a Gazebo-based simulator to support causal inference in HRSI scenarios, enabling the collection of observational data and allowing for interventions involving both the robot and humans.
The main limitation of our previous work was the inability to contextualise agent spatial behaviours, preventing the generation of stochastic and realistic human trajectories. Indeed, in ROSCausal\_HRISim, the agent's goal sequence is fully deterministic.
PeopleFlow represents its natural extension, and is designed for modelling human-human and human-robot spatial interactions in shared workspaces. Its key novelty lies in contextualising agent behaviours—both for humans and robots—based on scenario-specific factors such as time-dependent routines, robot states, and dynamic environmental conditions, all of which influence interaction patterns.
The simulator is built on Gazebo and ROS and features a TIAGo\footnote{\url{https://pal-robotics.com/robot/tiago/}\label{foot:TIAGo}} robot alongside multiple pedestrian agents. Human behaviour is modelled using the {\em pedsim\_ros}\footnote{\url{https://github.com/srl-freiburg/pedsim_ros}} library, which simulates group and individual movements based on a social force model~\citep{helbing1995social}. 
On top of this simulation stack, we introduce two new ROS nodes: the {\em Context Manager} and the {\em Context-Pedsim Bridge}.

The Context Manager node governs the scenario-specific contextual factors (e.g. time-dependent routines) and communicates the current context to the Context-Pedsim bridge node. The Context-Pedsim Bridge handles the generation of new tasks for pedestrian agents and computes their navigation plans accordingly. 
In more detail, we introduced a waypoint-based division of the environment into meaningful regions, such as workspaces, canteens, offices, and corridors. These waypoints serve both as semantic labels and as navigation nodes for guiding agent movement. Their connectivity is defined based on their spatial arrangement while ensuring that direct paths between them do not intersect obstacles or fall within the influence radius of other waypoints.
When an agent requests a new task, the Context-Pedsim Bridge node, based on the current context, selects a destination waypoint (i.e., an activity station) and assigns a random activity duration. The agent’s current position and assigned destination are then passed to an A\textsuperscript{*}-based path planner, which computes the shortest waypoint sequence to reach the goal. Once the agent reaches the destination, it remains stationary for the designated time to simulate engagement in the activity. Afterward, the task cycle restarts. The whole strategy is shown in Fig.~\ref{fig:simulator-approach}. Mapping destinations to specific waypoints and driving task assignment based on contextual rules (e.g., time of day), enables the production of more realistic and interpretable human activities.

Our simulator is designed to be modular, extensible and adaptable to different kinds of shared environments, contextual factors, and task structures. A Docker image of PeopleFlow is publicly available on GitHub\footref{foot:peopleflow}. 
\begin{figure}[t]
    \centering
    \begin{subfigure}{.475\textwidth}
        \includegraphics[trim={0cm 0cm 0cm 0.9cm}, clip, width=\textwidth]{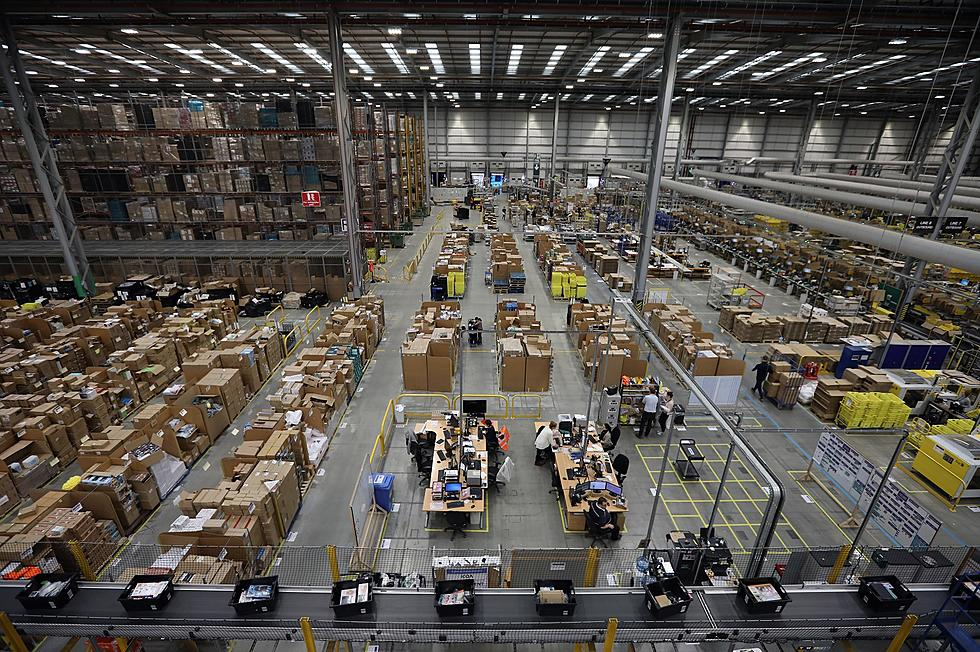}
        \caption{}\label{fig:simulator-examples-amazon}
    \end{subfigure}
    \begin{subfigure}{.475\textwidth}
        \includegraphics[trim={0.1cm 0cm 0.1cm 0cm}, clip, width=\textwidth]{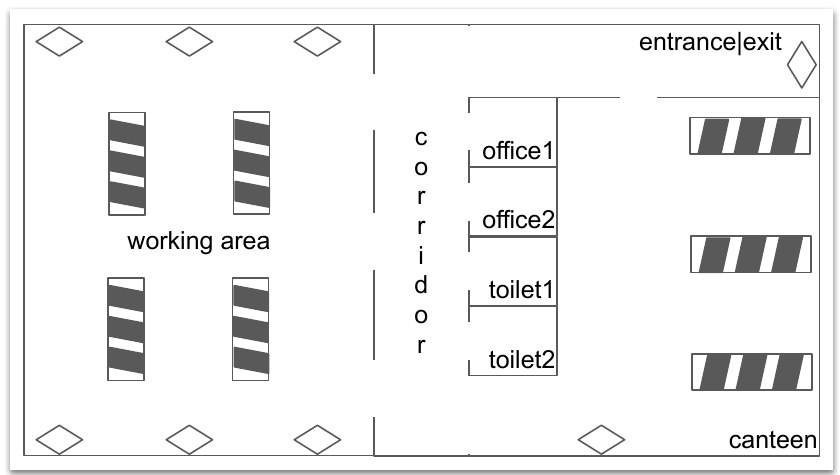}
        \caption{}\label{fig:simulator-examples-map}
    \end{subfigure}\\
    \begin{subfigure}{.475\textwidth}
        \includegraphics[trim={35.5cm 1.75cm 1.5cm 1.9cm}, clip, width=\textwidth]{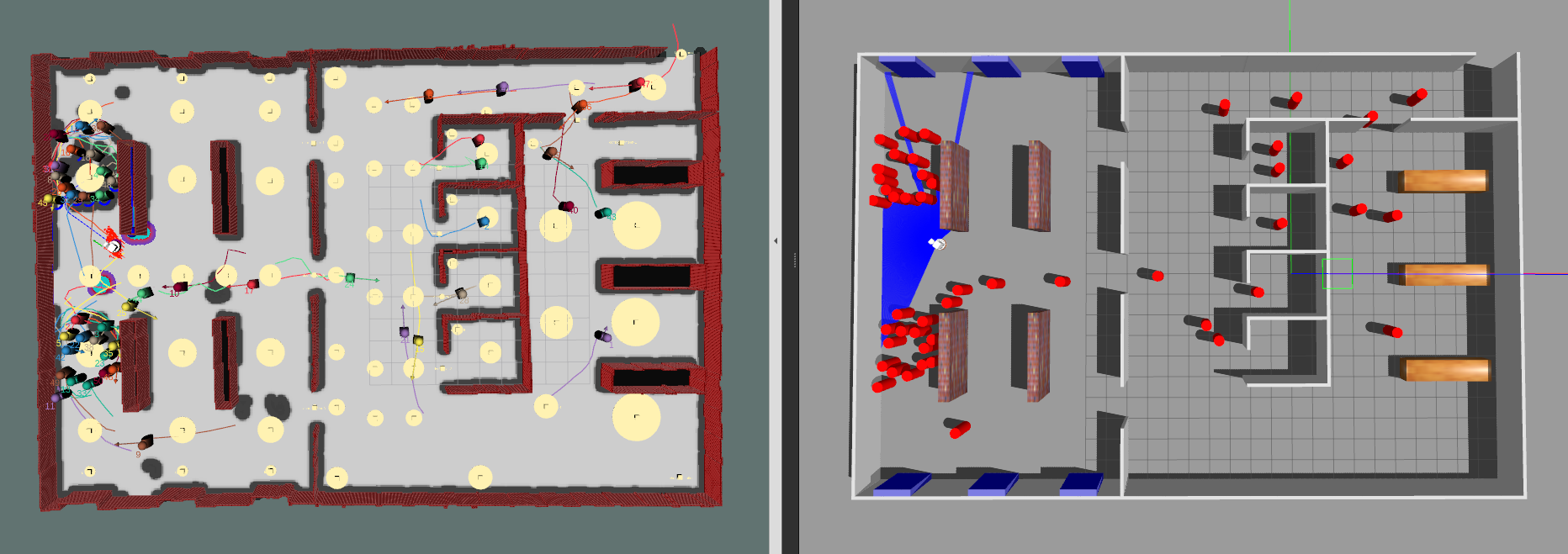}
        \caption{}\label{fig:simulator-examples-gazebo}
    \end{subfigure}
    \begin{subfigure}{.475\textwidth}
        \includegraphics[trim={1cm 1.5cm 35cm 1.5cm}, clip, width=\textwidth]{simulator-example-rviz-gazebo.png}
        \caption{}\label{fig:simulator-examples-rviz}
    \end{subfigure}
    \caption{(a) An illustrative example of an Amazon warehouse setting. (b) Warehouse-like scenario map with obstacles depicted as black-band rectangles, and robot target stations as diamonds. Its implementation in our simulator: (c) Gazebo view, (d) RViz view.}
    \label{fig:simulator-examples}
\end{figure}

\subsection{Context-driven Human-Robot Behaviours in Warehousing}\label{sec:simulator-scenario}
In this paper, we focus on a warehouse-like scenario--inspired by our funding project DARKO\footnote{\url{https://darko-project.eu/}\label{foot:darko}} and common industrial settings~\citep{dukic2012warehouse}--as a case study to evaluate our causality-based robot decision-making framework. Fig.~\ref{fig:simulator-examples} illustrates our approach: a typical Amazon warehouse scenario\footnote{\url{https://khak.com/amazon-facility-davenport}} (Fig.~\ref{fig:simulator-examples-amazon}) inspired the design of our simulated environment map (Fig.~\ref{fig:simulator-examples-map}). The environment consists of several main areas, including the entrance, two offices, two toilets, a canteen, a working area, and a corridor connecting all these spaces. Black-band rectangles indicate shelves in the working area where humans perform tasks and tables in the canteen, while diamonds represent target points for the robot (e.g., trays and conveyors). In Gazebo (Fig.~\ref{fig:simulator-examples-gazebo}), humans are represented as red cylinders, whereas in RViz (Fig.~\ref{fig:simulator-examples-rviz}), they are visualised as manikins. We now describe the contextual factors, robot features, and how we modelled worker activities within this environment.

\subsubsection{Contextual Factors}\label{sec:simulator-scenario-context}

We model key contextual factors that affect both human and robot behaviours, as well as their interaction in a warehouse scenario:
\begin{itemize}
    \item {\em Waypoint}~($W$), which identifies a specific area of the warehouse environment;
    \item {\em Time-slot}~($S$), which affects human activities and robot task execution, simulating different operational dynamics throughout a typical warehouse shift;
    \item {\em Robot charging status}~($C$), distinguishing between scenarios where the robot is actively navigating or stationary at a charging station;
    \item {\em Presence of static obstacle}~($O$), which models the presence of obstacles not included in the map of the environment (e.g., a forklift or a box left by a human). This factor affects robot velocity and battery consumption, as they require additional manoeuvrers.
\end{itemize}

\subsubsection{Robot Features}\label{sec:simulator-scenario-robot}
\begin{figure}[t]\centering
\includegraphics[trim={0cm 0cm 0cm 0cm}, clip, width=0.75\textwidth]{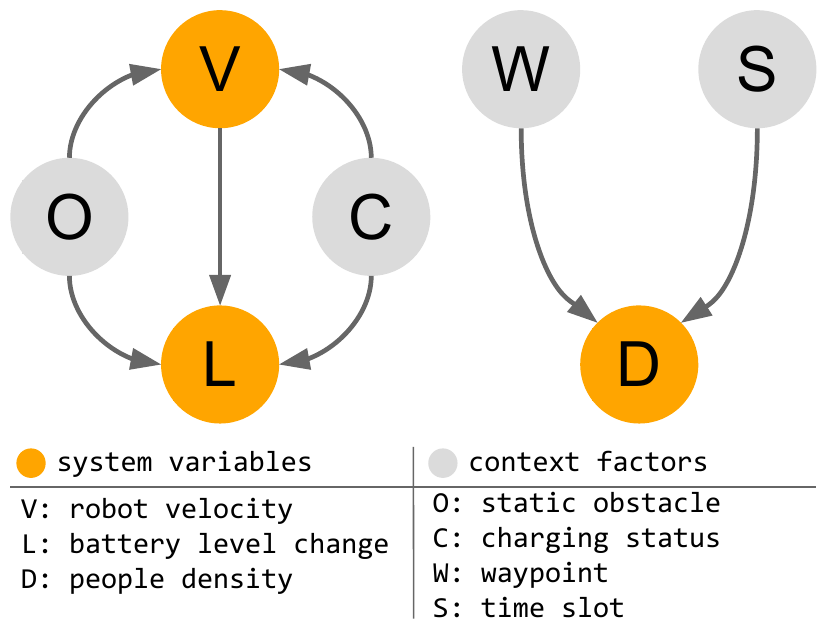}
\caption{Causal model of the scenario staged by PeopleFlow. Contextual factors $W$, $S$, $C$, and $O$ are shown in grey, while system variables $V$, $L$, and $D$ are highlighted in orange.}
\label{fig:simulator-hypothetical}
\end{figure}
Autonomy is a key factor for any mobile robot operating in real-world environments. Among the many aspects that influence autonomy, battery life is crucial, as it directly affects the robot's ability to perform tasks and activities.
In our simulator, we model battery dynamics, including both charging and discharging processes, to better approximate real-world conditions. Additionally, we model how the presence of an unexpected static obstacle in the environment affects the robot's battery dynamics and velocity. This allows us to more realistically simulate the robot’s operation in a warehouse environment, where battery consumption is a critical factor that must be considered in the decision-making process~\citep{tomy2019battery,tomy2020battery,fragapane2021planning,tosello2024opportunistic}.

We model the battery level's change ($L$) based on two contextual factors: the charging status ($C$) and the presence of an unexpected static obstacle ($O$). 
Inspired by previous works~\citep{hou2018energy}, the battery dynamic in discharging mode ($C = 0$) is based on two primary factors: {\em static} and {\em dynamic} consumption. Static battery consumption accounts for energy usage when the robot is idle, primarily due to active sensors and other essential components. In contrast, dynamic battery consumption represents the energy expended due to the robot’s movement. However, battery consumption increases in situations where the robot is forced to stop suddenly or perform manoeuvrers to avoid an unexpected static obstacle in the environment.
Instead, when the robot is in charging mode, the battery level increases at a predefined charging rate. To capture these factors, we model the change in battery level over time using the following equation:  
\begin{equation}\label{eq:delta-rb}
L_t =
\begin{cases}
-\Delta_t \Big( K_{s} + K_{d} \cdot V_t \Big) & C = 0, O = 0\\
-\Delta_t \Big( K_{s} + K_{d} \cdot V_t \Big) K_{o} & C = 0, O = 1\\
+\Delta_t \cdot K_{c} & C = 1
\end{cases}
\end{equation}
where $L_t$ represents the battery level's change at time step $t$, $\Delta_t$ is the time step duration, $K_s$ is a constant representing static battery consumption,  
$K_d$ is a coefficient related to battery consumption due to motion, $V_t$ is the robot’s velocity at time step $t$, $K_o$ is a coefficient representing the increase in battery consumption caused by static obstacles that force the robot to modify its behaviour unexpectedly. $K_c$ is the charging rate when the robot is in charging mode. The two contextual factors, $C$ and $O$, indicate the robot's charging status and the presence of a static obstacle, respectively. $C$ is $0$ when the robot is discharging and $1$ when it is charging, while $O$ is $0$ when no obstacle is present and $1$ when an obstacle is detected.

The $K_s$ and $K_d$ parameters were calibrated to be consistent with the official TIAGo robot's documentation\footnote{TIAGo datasheet:~\url{https://static1.squarespace.com/static/5ab0712975f9ee24a7be79df/t/5ba9cc4f652deac39f368c37/1537854552932/Datasheet_TIAGo-Hardware-Software.pdf}}, which states a typical battery duration of $4$ to $5$ hours. We used these manufacturer specifications to derive the parameters as follows:
\begin{itemize}
    \item Static consumption $K_s$: We assumed the $5$-hour ($5$~h) duration represents the robot in an idle (static) state, where sensors and systems are active but the motors are not moving. $K_s$ was thus calculated as the constant drain rate per second required to deplete $100\%$ of the battery in 5 hours: $K_s = 100\%/(5~\text{h} \times 3600~\text{s/h}) \approx 0.0056 \%/\text{s}$
    \item Dynamic Consumption $K_d$: We assumed the $4$-hour duration represents continuous operation at maximum velocity, $V_{\text{max}}$. We calculated $K_d$ by taking the total drain rate for the 4-hour duration ($100\% / (4~\text{h} \times 3600~\text{s/h})$), subtracting the static drain $K_s$, and then dividing by $V_{\text{max}}$: $K_d \approx 0.0027 \%/\text{m}$.
    \item Obstacle penalty $K_o$: This parameter is not specified in the official documentation, as it models the non-standard cost of unexpected manoeuvrers (e.g., sudden braking). This coefficient was set empirically to \textbf{3.0}, representing an increase in consumption during such events.
\end{itemize}

Thus, the battery level at time $t$ is updated as follows:
\begin{equation}
B_t = B_{t-1} + L_{t-1}
\end{equation}
where $B_t$ and $B_{t-1}$ denote the battery percentage at time $t$ and $t-1$ respectively, and $L_{t-1}$ represents the battery level's change, as modelled by~(\ref{eq:delta-rb}), at time $t-1$.

By incorporating battery dynamics into the simulation, we account for the impact of its consumption and recharging needs on the robot’s operation. This enables more realistic decision-making, such as determining whether to continue or abort a task based on battery levels, planning charging station visits, and selecting a navigation path that balances safety, time efficiency, and battery consumption. 

The expected causal relationships between the mentioned factors derive from the model in (\ref{eq:delta-rb}) and are as follows:
\begin{itemize}
    \item $L \xleftarrow{} O \xrightarrow{} V$: $L$ is affected by $O$. Moreover, $O$ influences $V$ because the robot is forced to stop or maintain a low velocity when encountering an unexpected obstacle.   
    \item $L \xleftarrow{} C \xrightarrow{} V$: When $C = 0$, $L$ is discharging, whereas when $C = 1$, $L$ is charging. Furthermore, $C = 1$ also forces the robot to remain stationary at the charging station, meaning $V = 0$.  
    \item $V \xrightarrow{} L$: The robot's velocity affects the battery change $L$ in discharging mode.  
\end{itemize}
Although we use time-series data, we omit explicit time indices $t$ and $t-1$ for simplicity, as all the described links represent contemporaneous relationships occurring at time $t$. The considered relationships are illustrated by the Directed Acyclic Graph~(DAG) in Fig.~\ref{fig:simulator-hypothetical}.

\subsubsection{Worker Activities}\label{sec:simulator-scenario-worker}
In a warehouse, workers are assigned specific tasks at different times of the day. The activities assigned to each person are generally the same every day~\citep{meneweger2018factory}.  
For instance, warehouse operators may be responsible for unloading goods in the morning, sorting and shelving items during the day, and preparing shipments in the evening. These tasks are generally repeated every working day. As a result, certain areas of the warehouse experience higher worker traffic during specific time periods, creating congestion zones.

In order to model this phenomenon, we divided the environment into distinct areas (waypoints) and introduced time as a contextual factor to differentiate workers' activities at different times of the day. Each waypoint represents a specific area of the warehouse that experiences varying levels of congestion depending on the time, making it an essential contextual factor in our model.  

To capture the temporal dynamics of workers activity, we divided the working period into hourly time-slots from 08:00 to 18:00, along with an additional off period. Within each time-slot, worker activities are predefined and consistent, reflecting the structured nature of warehouse operations.  
The 08:00–09:00 slot models workers' arrival and the beginning of the working day. To enhance realism, we model each worker arriving at a random time within this slot before starting their assigned activity. Time-slots from 09:00 to 13:00 represent the normal morning working period, during which workers are typically present in operational areas such as shelves and offices. The 13:00–14:00 slot represents the lunch break, where workers are primarily in the canteen area.  
Following this, time-slots from 14:00 to 18:00 model the afternoon working period, with the 17:00–18:00 slot representing the end of the working day, when workers finish their tasks and leave the warehouse after completing their standard eight-hour shift. After 18:00, we model the off period, during which no workers are present in the warehouse.

\begin{figure}[t]
    \centering
    \begin{subfigure}{0.54\textwidth}
        \includegraphics[trim={0cm 0.3cm 3cm 1.2cm}, clip, width=\textwidth]{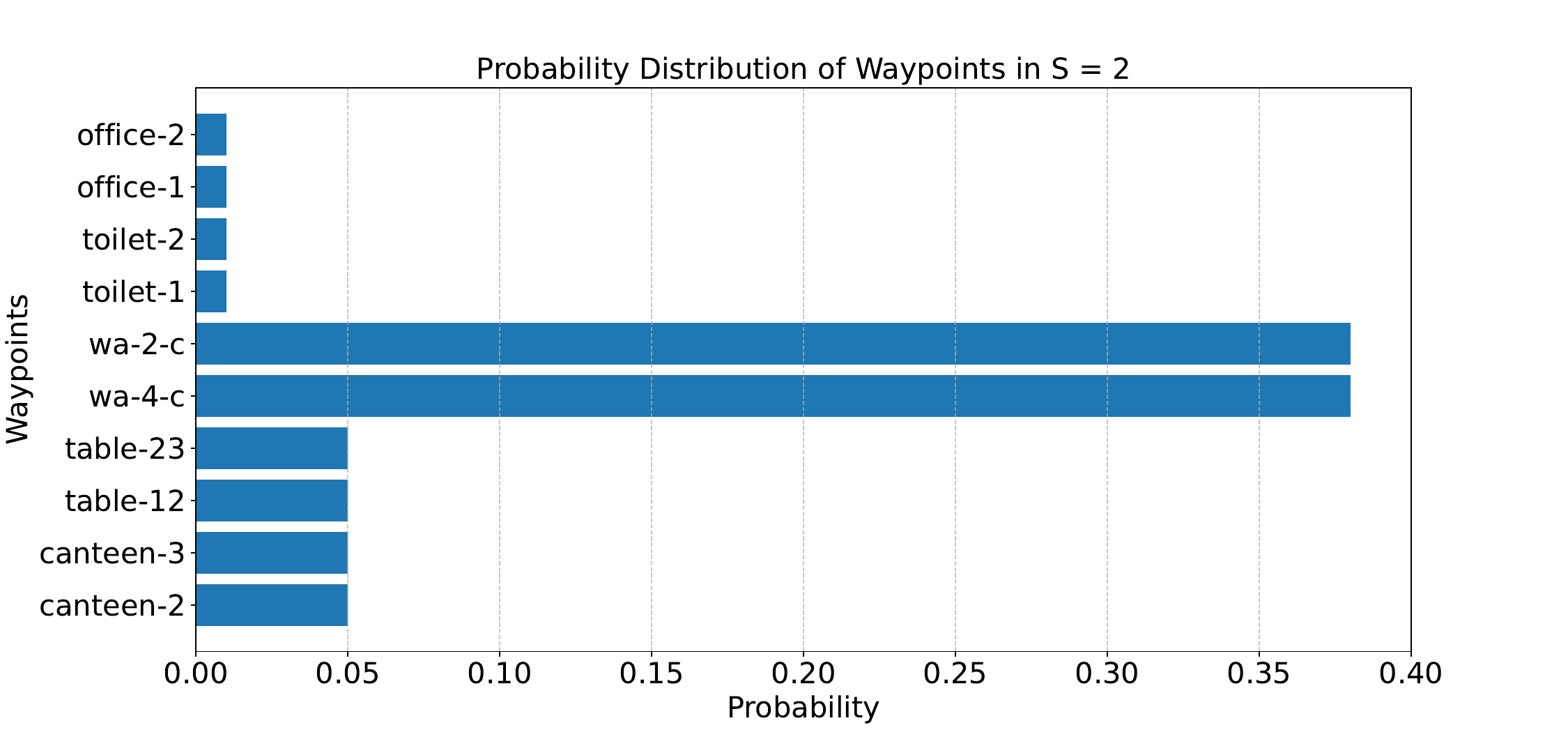}
        \caption{}\label{fig:simulator-pd-pdh2}
    \end{subfigure}
    \begin{subfigure}{0.45\textwidth}
        \includegraphics[trim={3cm 0cm 0cm 0.5cm}, clip, width=\textwidth]{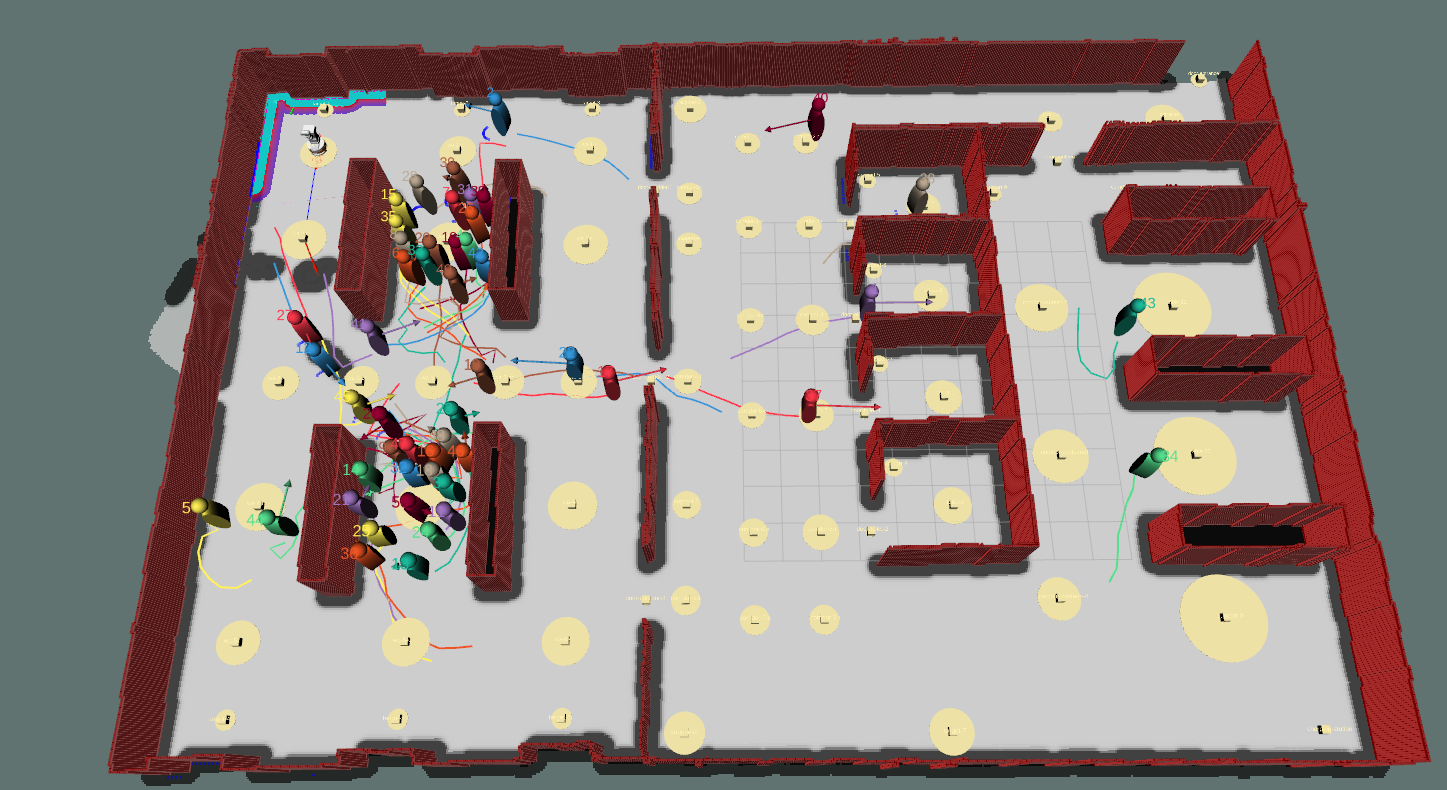}
        \caption{}\label{fig:simulator-pd-h2}
    \end{subfigure}\\
    \begin{subfigure}{0.54\textwidth}
        \includegraphics[trim={0cm 0.3cm 3cm 1.2cm}, clip, width=\textwidth]{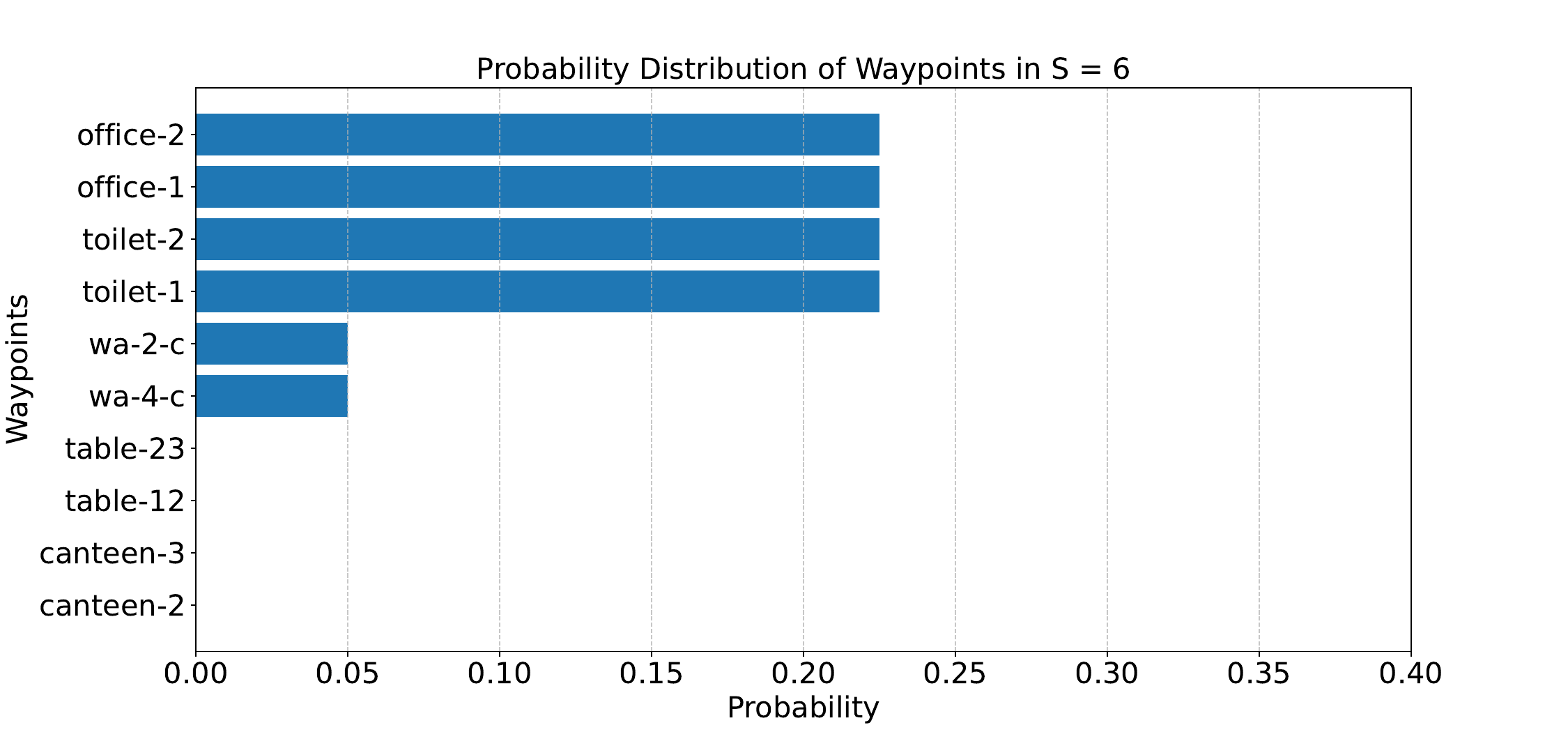}
        \caption{}\label{fig:simulator-pd-pdh6}
    \end{subfigure}
    \begin{subfigure}{0.45\textwidth}
        \includegraphics[trim={3cm 0cm 0cm 0.5cm}, clip, width=\textwidth]{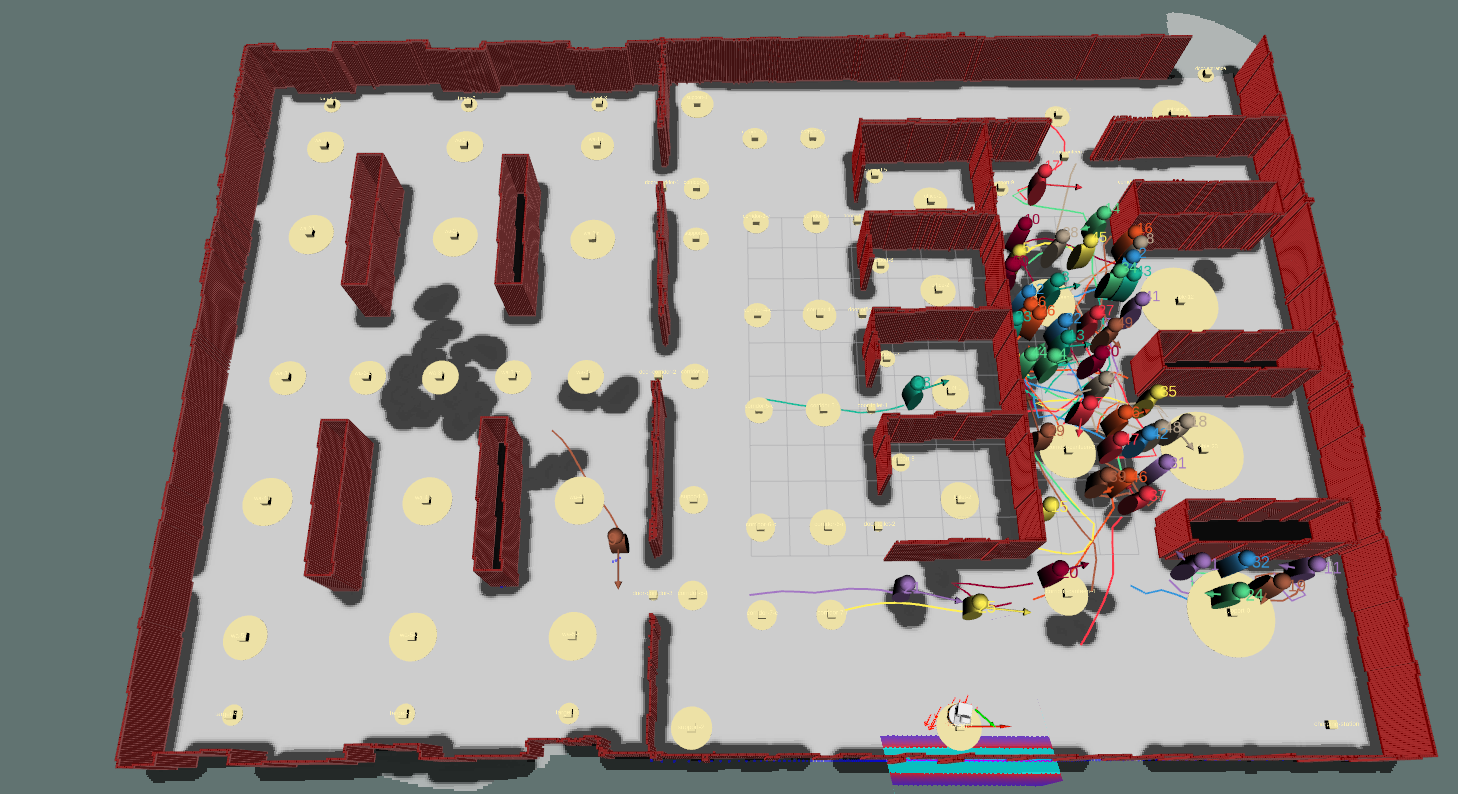}
        \caption{}\label{fig:simulator-pd-h6}
    \end{subfigure}
    \caption{Two examples of people congestion areas at different time-slots. (a) Probability distribution of waypoint choices for people in time-slot $2$ (09:00–10:00). (b) RViz view showing people primarily concentrated at the waypoints with the highest probabilities from the distribution graph. (c) Probability distribution of waypoint choices for people during lunchtime ($S6$, 13:00–14:00). (d) RViz view showing the majority of workers gathered in the canteen area.}\label{fig:simulator-pd}
\end{figure}
An example of a congestion zone in two different time-slots is illustrated in Fig.~\ref{fig:simulator-pd}. 
Based on this, we defined a variable called {\em people density} ($D$) to quantify the number of people in a specific waypoint. For each waypoint $W$ in the set of waypoints $\Omega$, the density $D$ is defined as follows:  
\begin{equation}\label{eq:pd}
    D = \frac{\text{Number of people in }W}{\pi \cdot r_{W}^2}
\end{equation}
where $r_{W}$ is the radius of the waypoint $W$.

By incorporating both waypoints and time-slots as contextual factors, our model effectively represents fluctuating congestion levels within the warehouse. This enables a more accurate simulation of human-robot interactions, where robot navigation decisions account for dynamically changing environmental conditions. 
Note that while the time-slot serves as the primary contextual driver for worker activity, the model is not deterministic. Indeed, each time-slot is associated with a probability distribution over the possible waypoints, not a fixed assignment, which in turn introduces stochasticity. This allows workers to occasionally select less probable destinations, simulating natural variations in behaviour such as taking unscheduled breaks or moving between different areas, as shown in Figs.~\ref{fig:simulator-pd-pdh2}~and~\ref{fig:simulator-pd-h2}. In the example, you can notice that, although it is a ``working time'' time-slot where the working area has the highest probability of being occupied, a number of workers can still be observed in the canteen and corridors, demonstrating the non-deterministic nature of the simulation.
Following this, the causal influence from the contextual factors time-slot $S$ and waypoint $W$ on the people density is $W \xrightarrow{} D \xleftarrow{} S$. Again, for simplicity, we have omitted the explicit time indices $t$ and $t-1$. The latter causal relations are added to the model in Fig.~\ref{fig:simulator-hypothetical}.

\subsection{Causality-based Robot Decision Making in PeopleFlow}\label{sec:simulator-causal}
Our causality-based decision making framework presented in Sec.~\ref{sec:causal-inference-inference} is a general causal inference approach adaptable to various scenarios. Here, we provide more details on the factors that we estimate using causal inference and how they are used in the A\textsuperscript{*} algorithm to decide whether to execute or not a given task, and in case to determine the best path.

\paragraph{\textbf{Causal Inference}}
Given the robot and workers features extracted from the simulator, as well as the context factors explained in Sec.~\ref{sec:simulator-scenario-context}, the crucial aspects affecting robot task completion are the robot battery consumption~$L$ and the people density~$D$ in specific areas of the environment. Given their dependencies with other variables and/or contextual factors, the robot can infer $L$ based on its velocity $V$ and charging status $C$, while $D$ can be estimated using the specific time-slot $S$ and waypoint $W$. 
These inferences are made under the assumption that the robot has knowledge of its own velocity and charging status. Additionally, in a warehouse shared with workers, we assume that the robot is aware of the current time, enabling it to determine $S$, as well as the predefined waypoints in the environment. This allows for a probabilistic prediction of congested areas based on context (time and location), rather than a simple reactive forecast based only on past density values.

For instance, consider a scenario where the robot is assigned the task ``transport an object to the delivery point,'' as depicted in Fig.~\ref{fig:causal-inference}. To determine in advance whether the task can be successfully completed or if it should be immediately refused, the robot poses the following query:  
\begin{center}  
    {\em ``What if I go to the delivery point now at velocity $v$?''}  
\end{center}  
This query involves two causal interventions: one on the time (``now'') and another on the velocity ($v$). These interventions are essential for estimating $L$ and $D$, allowing the robot to make an informed decision about whether to proceed with the task. Specifically, the robot can query the causal inference engine to estimate the following probability distributions: %
\begin{equation}\label{eq:causal-inference-delta-rb}
    \text{P}(L = l \mid do(V = v), C = c) \quad \forall l \in \Gamma
\end{equation}
\begin{equation}\label{eq:causal-inference-pd}
    \text{P}(D = d \mid do(S = s), W = w) \quad \forall d \in \Psi, \forall w \in \Omega
\end{equation}
where $\Gamma$ and $\Psi$ denote the sets of possible discretised values for $L$ and $D$, respectively, and $\Omega$ is the set of waypoints. Eqs.~(\ref{eq:causal-inference-delta-rb})~and~(\ref{eq:causal-inference-pd}) return the probability distribution of $L$ and $D$, respectively, given the intervention and conditioning variables. These probability distributions are then used to retrieve the expected $L$ and $D$ as follows:
\begin{equation}\label{eq:causal-inference-delta-rb-exp}
    \hat{L} = \mathbb{E}[L \mid do(V = v), C = c]
\end{equation}
\begin{equation}\label{eq:causal-inference-pd-exp}
    \hat{D} = \mathbb{E}[D \mid do(S = s), W = w] \quad \forall w \in \Omega
\end{equation}
To clarify, the robot queries the causal inference engine to determine whether it can start the task at the current $S$ and travel at velocity $v$. Hence, for each waypoint in $\Omega$, these two interventions allow the causal inference engine to estimate the expected values of $L$ and $D$, which represent the cost associated with the specific waypoint in terms of battery consumption and the risk of encountering people or getting stuck, respectively. 

It is important to note the different nature of these two queries. The estimation of people density, $\text{P}(D = d \mid do(S = s), W = w)$, can be simplified to a standard conditional probability, $\text{P}(D = d \mid S = s, W = w)$, because there is no confounding variable between time-slot ($S$) and people density ($D$) in our model (see Fig.~\ref{fig:simulator-hypothetical}). In contrast, estimating battery consumption ($L$) requires a causal approach. The relationship between velocity ($V$) and battery level ($L$) is confounded by the presence of unexpected obstacles ($O$). Since the robot does not know in advance if it will encounter an obstacle, it cannot simply condition on $O$. Therefore, a proper causal adjustment using do-calculus is essential, motivating our choice of a causal inference engine over a standard Bayesian method.

These estimates are then used as cost components for navigation planning.
In particular, the battery cost for traversing an arc between two waypoints is first computed by determining the traversal time, which is the distance between the waypoints divided by the velocity~$v$ (used in the intervention). This duration is then multiplied by the expected $\hat{L}$ from Eq.~\eqref{eq:causal-inference-delta-rb-exp}, which corresponds to the battery consumption per time step when moving at velocity~$v$. The result is the expected battery cost for that arc.
These estimated values, along with the distance between waypoints, are finally used in a heuristic function defined below.

\paragraph{\textbf{Heuristic}} Given a path $\pi = \{w_1, w_2, ..., w_n\}$ from the starting waypoint $w_1$ to the destination $w_n$, the A\textsuperscript{*} algorithm evaluates paths using the standard cost function:
\begin{equation}\label{eq:causal-inference-A*}
f(w_i) = g(w_i) + h(w_i)
\end{equation}
where $g(w_i)$ represents the cumulative cost from the start node $w_1$ to the current node $w_i$, while $h(w_i)$ is the heuristic estimate of the remaining cost from $w_i$ to $w_n$. For our problem, we define the heuristic cost function as:
\begin{equation}\label{eq:causal-inference-heuristic}
    h(w_i) =
    \sum\limits_{i=1}^{n-1} \Big(\lambda_\delta \cdot \delta(w_i, w_{i+1}) + \lambda_{D} \cdot \hat{D}(w_i) + \lambda_{L} \cdot |\hat{L}(w_i, w_{i+1})|\Big)
\end{equation}
where $\delta(w_i, w_{i+1})$ represents the Euclidean distance between two consecutive waypoints, $\hat{D}(w_i)$ is the expected people density at waypoint $w_i$ from the causal inference engine, $\hat{L}(w_i, w_{i+1})$ is the expected battery consumption to reach waypoint $w_{i+1}$ from $w_i$, and $\lambda_\delta, \lambda_{D}, \lambda_{L}$ are weighting coefficients that balance the trade-off between distance minimisation, congestion avoidance, and battery efficiency.

\paragraph{\textbf{Decision Making}} As already mentioned, we want the causal inference engine to be able not only to choose the path but also to determine whether the task should be executed. This decision must be based on the cumulative battery cost along the path $\pi$, defined as:
\begin{equation}\label{eq:total-battery-consumption}
    C_{L}(\pi) = \sum_{i=1}^{n-1} |L(w_i,w_{i+1})|
\end{equation}
and a minimum battery threshold ($B_{\text{min}}$), a common safeguard in robotics to prevent critical failures. If the battery level falls below this threshold, the robot typically returns to the charging station or deactivates some power-consuming services. For instance, the TIAGo robot disables its arm motors, making manipulation tasks not possible.
More specifically, if the remaining battery $B_t - C_{L}(\pi)$ after a candidate path $\pi$ would fall below a threshold $B_{\text{min}}$, the path $\pi$ must be discarded. 
This heuristic function~(\ref{eq:causal-inference-heuristic}) is integrated into Eq.~\ref{eq:causal-inference-A*} and guides the A\textsuperscript{*} algorithm to expand paths from the starting waypoint to the destination, while minimising the total estimated cost along the path $\pi$. The heuristic function ensures that selected paths balance distance, congestion avoidance, and battery efficiency within the minimum battery constraint.
If the A\textsuperscript{*} finds a path that satisfies such constraint, the decision block returns it to the robot for execution. Otherwise, an abort signal is issued.

It is important to note that this decision is made once at the start of the task. We do not re-check or re-plan using the causal model during execution. This was a deliberate design choice for two reasons:
\begin{itemize}
    \item Crowd density is a relatively slow-moving variable. Therefore, if an area is inferred to be busy at the start of a task, it is likely to remain busy for the task's duration. A single pre-task check is sufficient.
    \item For dynamic events, such as encountering an unexpected static obstacle, we use a two-level system: ({\em i}) before the task, our battery consumption query (Eq.~\ref{eq:causal-inference-delta-rb}) explicitly accounts for the possibility of an obstacle. The do-calculus adjusts for the confounder $O$, so the resulting battery estimate $\hat{L}$ already considers the average cost of potential avoidance; ({\em ii}) during the task, if the robot physically encounters an obstacle, the standard ROS local planner handles the real-time avoidance.
\end{itemize}
This strategy provides a robust compromise: we make an informed, battery-aware strategic decision at the start, without incurring the high computational cost of re-running the causal query loop during execution.

\section{Experiments}\label{sec:exp}
\begin{figure}[t]\centering
\includegraphics[trim={0cm 2.2cm 0cm 0.2cm}, clip, width=\textwidth]{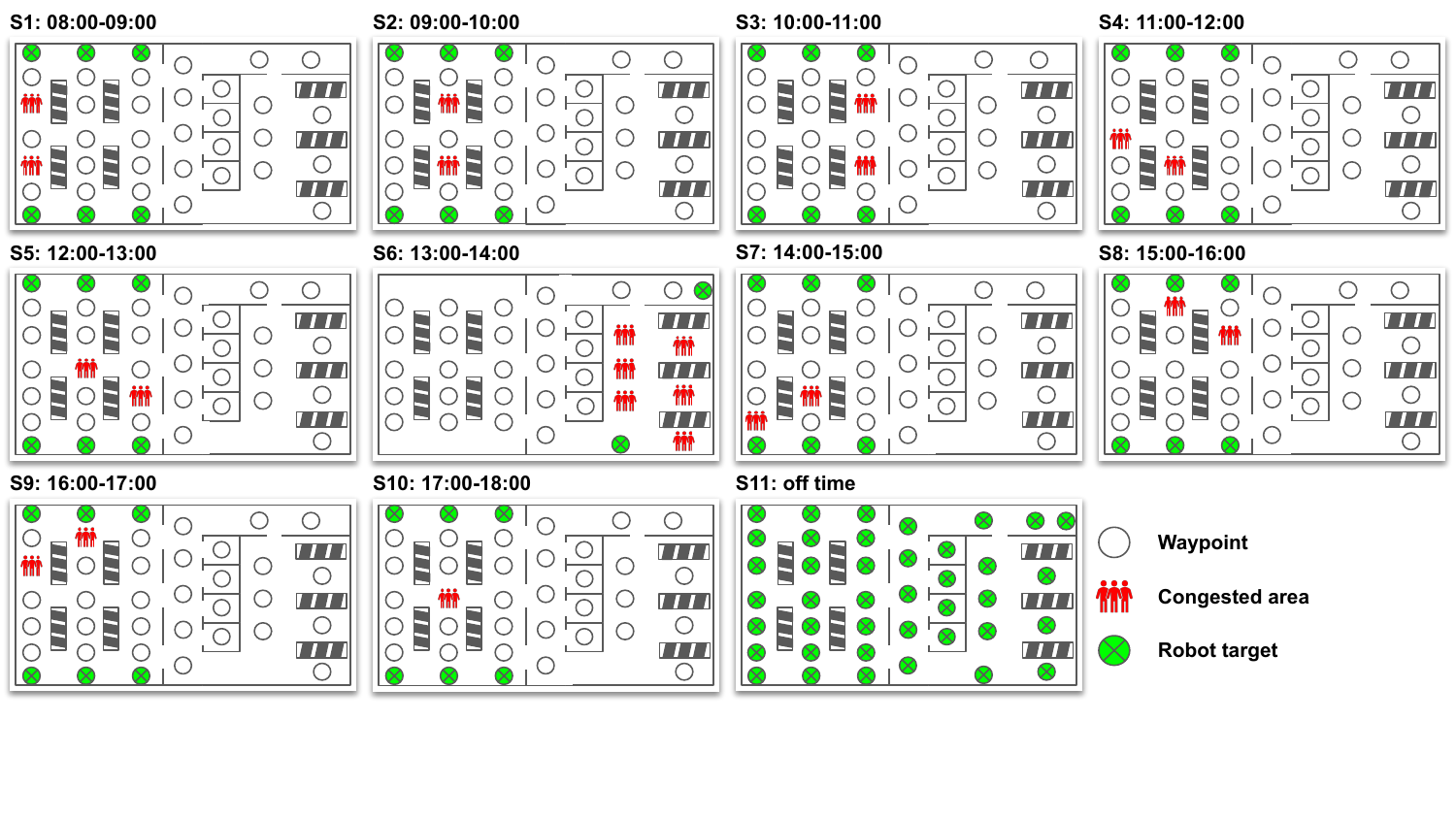}
\caption{Experimental scenario simulated using our simulator, as described in Sec.~\ref{sec:simulator}. The scenario is divided into 11 time-slots~($S$), during which worker congestion areas and robot tasks change dynamically.
}
\label{fig:exp-scenario}
\end{figure}
This section evaluates the proposed causality-based decision-making framework (Sec.~\ref{sec:causal-inference}) to demonstrate its benefits for robot efficiency and safety in a dynamic, shared environment.

Our framework's core contribution is twofold: ({\em i}) it uses a causal model to improve path selection by scoring routes based on expected congestion and battery drain, and ({\em ii}) preemptively refuse tasks that are likely to fail due to low battery.
To precisely quantify the individual and combined impact of these two mechanisms, our evaluation is structured as an ablation study. We compare the performance of four distinct approaches:
\begin{itemize}
    \item Baseline (Shortest-Path): a standard A\textsuperscript{*} planner using only Euclidean distance, which must accept all tasks;
    \item Causal Routing: our causal heuristic (Eq.~\ref{eq:causal-inference-heuristic}) for path selection, but forced to accept all tasks;
    \item Refusal-Only: the baseline planner, but with our causal task refusal mechanism enabled;
    \item Full Causal Framework: the complete system, using both causal routing and causal refusal.
\end{itemize}
This comparative analysis, conducted within our PeopleFlow simulator, clearly isolate the gains from smarter routing versus intelligent task refusal, validating the overall contribution to safety and efficiency.

\subsection{Experimental Setup}\label{sec:exp-learning}
All evaluation experiments were conducted on a desktop computer (12th Gen Intel Core i9-12900F × 24, NVIDIA GeForce RTX 4070, 32 GB RAM), based on Ubuntu 20.04~LTS and ROS-Noetic.

Collecting real-world data in a warehouse-like environment with multiple workers and dynamic contextual factors would require significant time and financial resources. Given these constraints, we evaluate our decision-making framework in simulation, similarly to other recent works on multi-agent systems and human-robot social navigation~\citep{mavrogiannis2019multi, mavrogiannis2021hamiltonian}. Our simulator enables the creation of different controlled and repeatable HRSI scenarios, allowing for a systematic analysis of how environmental conditions, agent behaviours, and task constraints influence HRSIs. The data generated by multiple simulated scenarios is processed by our framework to learn a causal model. The latter is finally used to perform actual inference and decision-making, as explained in the following sections.

\paragraph{\textbf{Simulated Scenarios}}
We designed 11 different scenarios with varying worker congestion areas and robot tasks, as illustrated in Fig.~\ref{fig:exp-scenario}. These scenarios correspond to 11 different time-slots ($S1$-$S11$) throughout the day, capturing realistic variations in worker activity.
At each time-slot, we assigned a higher probability for waypoints ($W\!s$) to be occupied by workers. This models the fact that, at different times of the day, workers perform specific tasks in designated areas. For instance, in $S 1$–$S 5$ and $S 7$–$S 10$, workers are engaged in normal duties, primarily occupying working areas such as shelves and offices. The slot $S 6$ represents lunchtime, during which workers are concentrated in the canteen area. Finally, $S 11$ models the off-time period when workers are typically absent.
Some congestion areas (represented by a red crowd icon in Fig.~\ref{fig:exp-scenario}) correspond to waypoints with the highest probability of being occupied. However, workers are still allowed to move to other locations with lower probabilities, i.e., non-congested areas represented by white circles in Fig.~\ref{fig:exp-scenario}. This models the natural behaviour of workers taking breaks, going to the toilet, visiting the canteen, or occasionally moving to other locations beyond their primary working area.

Also the robot's tasks vary depending on the time-slot to align with human activities.
In $S 1$–$S 10$ (except $S 6$), the robot continuously moves between designated targets (represented as green crossed circles in Fig.~\ref{fig:exp-scenario}) at the top and bottom of the working area. Each time, it randomly selects one target from the top and one from the bottom, simulating a pick-and-place task.
In $S 6$ (lunchtime), the robot assists in the canteen by performing service-related tasks, such as carrying trays or collecting and disposing of waste in bins. The robot's goals in Fig.~\ref{fig:exp-scenario} ($S 6$) reflect this scenario: one near the canteen and the other near the entrance.
Finally, in $S 11$, when workers are no longer in the warehouse, the robot is assigned a cleaning task. During this period, it navigates through all the waypoints, covering the entire workspace. This final time-slot serves to validate the behaviour of our decision-making approach in an empty environment, ensuring that it performs identically to the baseline when no human-robot interactions occur.

\paragraph{\textbf{Data Generation}}
\begin{figure}[!t]
    \centering
    \includegraphics[trim={14cm 2.5cm 8cm 4.5cm}, clip, width=0.75\textwidth]{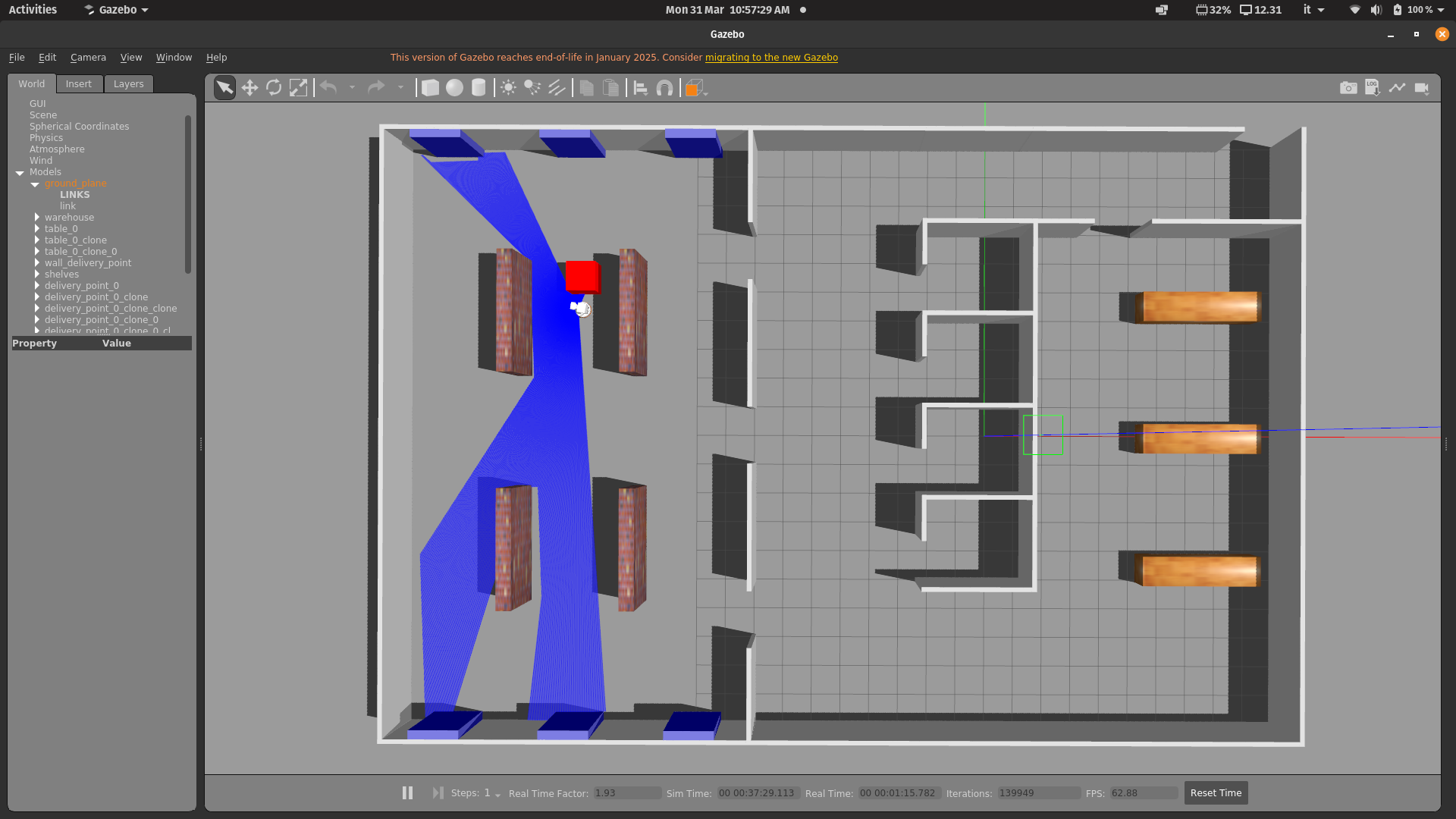}
    \caption{Gazebo view of our simulated warehouse-like scenario staged by PeopleFlow. The scenario shows workers (red cylinders) performing tasks in the working area, as well as in the offices, corridor, and canteen. The TIAGo robot navigates between target positions to complete tasks while avoiding unexpected obstacles (purple cubes) in the environment.}
    \label{fig:exp-scenario-obstacle}
\end{figure}
We collected a full-day simulated data, including robot’s velocity $V$, battery level $B$, and number of people at each waypoint $W$. 
Additionally, contextual factors such as the presence of unexpected obstacles $O$ (modelled by purple boxes as shown in Fig.~\ref{fig:exp-scenario-obstacle}), charging status~$C$, waypoint~$W$, and time-slot~$S$ were directly extracted from the simulator.
We simulated the presence of 50 workers in the warehouse-like scenario, ensuring that their goal selection aligned with the scenario shown in Fig.~\ref{fig:exp-scenario}.

We set the obstacle appearance probability to a value of $25\%$, representing a trade-off between completely obstacle-free environments and highly cluttered ones (e.g., $50\%$ probability). While the latter represents a more challenging scenario, it is unlikely to occur in typical real-world settings. This setup reflects realistic conditions where workers occasionally leave objects--such as boxes, pallets, or forklifts--in corridors or workspaces, thereby affecting the robot’s navigation, velocity, and battery discharge dynamics.

We assumed the robot starts with a fully-charged battery and set its parameters to be consistent with its actual lifetime\footref{foot:TIAGo} in both idle and motion modes.
During this data collection phase, the robot navigated the environment using a ROS-based navigation stack that relies on the A\textsuperscript{*} algorithm as global planner~\citep{marder2010office,zheng2023navigation}, employing the shortest path as its heuristic to reach its goal. This shortest-path strategy serves as a baseline for evaluating our causal decision-making approach, which will be analysed in the following sections.

\paragraph{\textbf{Data Post-processing}}
To reduce the dataset size and prepare it for our framework's learning pipeline, we first subsampled the recorded data according to the Nyquist–Shannon sampling theorem. We then post-processed the data to derive the necessary variables: the battery consumption variable $L$ was computed from the recorded battery level $B$, while people density $D$ was derived using Eq.~(\ref{eq:pd}), based on the number of people at each waypoint. These transformations allowed us to construct the full set of variables required for causal model learning. Finally, to enable parameter learning, we discretised the post-processed data using the pyAgrum discretiser.

\paragraph{\textbf{Causal Model Learning}}
The data post-processed in the previous step is then used for causal discovery. We assume an underlying first-order Markov process, and therefore consider a one-step time lag in the causal discovery stage. This Markovian assumption is widely used in the literature for modelling and predicting human spatial behaviours and human-robot collaboration~\citep{asahara2011pedestrian,liu2017human,1400974}. We use the causal discovery algorithm J-PCMCI\textsuperscript{+} to handle context variables. As stated in Sec.~\ref{sec:causal-inference-learning}, J-PCMCI\textsuperscript{+} is specifically designed to model context variables (like $S$, $W$, $C$, $O$) as external factors that affect the system (like $D$, $V$, $L$) without being influenced by it. This structural constraint inherently enforces ``expected'' real-world directionality by only permitting causal links from context-to-system and system-to-system, while disallowing links from system-to-context (e.g., a time-slot can cause cafeteria congestion, but cafeteria congestion cannot change the time-slot). Specifically, for the conditional independence tests of J-PCMCI\textsuperscript{+}, we employ the RegressionCI test (from the \texttt{tigramite} package\footnote{\url{https://github.com/jakobrunge/tigramite}}) given its ability to handle a mixed dataset of continuous (e.g., velocity) and discrete/categorical (e.g., time-slot) variables, as in our case. We use the standard significance threshold of $\alpha = 0.05$ for these tests.

The data post-processed in the previous step is then used for causal discovery. We assume an underlying first-order Markov process, and therefore consider a one-step time lag in the causal discovery stage. This Markovian assumption is widely used in the literature for modelling and predicting human spatial behaviours and human-robot collaboration~\citep{1400974,asahara2011pedestrian,liu2017human}. We use J-PCMCI\textsuperscript{+} to handle context variables. As stated in Sec.~\ref{sec:causal-inference-learning}, J-PCMCI\textsuperscript{+} is specifically designed to model context variables (like $S$, $W$, $C$, $O$) as external factors that influence the system (like $D$, $V$, $L$) without being influenced by it. This structural constraint inherently enforces ``obvious'' real-world directionality by only permitting causal links from context-to-system and system-to-system, while disallowing links from system-to-context (e.g., a time-slot can cause cafeteria congestion, but cafeteria congestion cannot change the time-slot). Specifically, for the conditional independence tests, we employ the RegressionCI test (from the \texttt{tigramite} package\footnote{\url{https://github.com/jakobrunge/tigramite}}) given its ability to handle a mixed dataset of continuous (e.g., velocity) and discrete/categorical (e.g., time-slot) variables, as in our case. We use the standard significance threshold of $\alpha = 0.05$ for these tests.

Note that the J-PCMCI\textsuperscript{+} algorithm is executed in a single run on the full simulated dataset, as analysing the entire working day was essential to capture all contextual dynamics. This aspect would be missed if smaller data windows were selected for the causal discovery. A detailed consistency analysis of the causal model across different data windows is outside the scope of this paper, whose focus is on the novel end-to-end decision-making framework. However, a detailed analysis of the causal discovery for various data parameters---i.e., time-series length and sampling frequency---was presented in our prior work on ROS-Causal~\citep{castri2024roman}.

In terms of computational cost, it is well-established that causal discovery scales with the number of variables~\citep{runge_causal_2018,castri2023enhancing,castri2024candoit}. For our specific model, which has 7 variables, the J-PCMCI+ algorithm took 4 minutes and 12 seconds to run on the hardware specified in Sec.~5.

\begin{figure}[t]
\centering
\includegraphics[trim={3.5cm 3.9cm 2.7cm 3.8cm}, clip, width=0.75\textwidth]{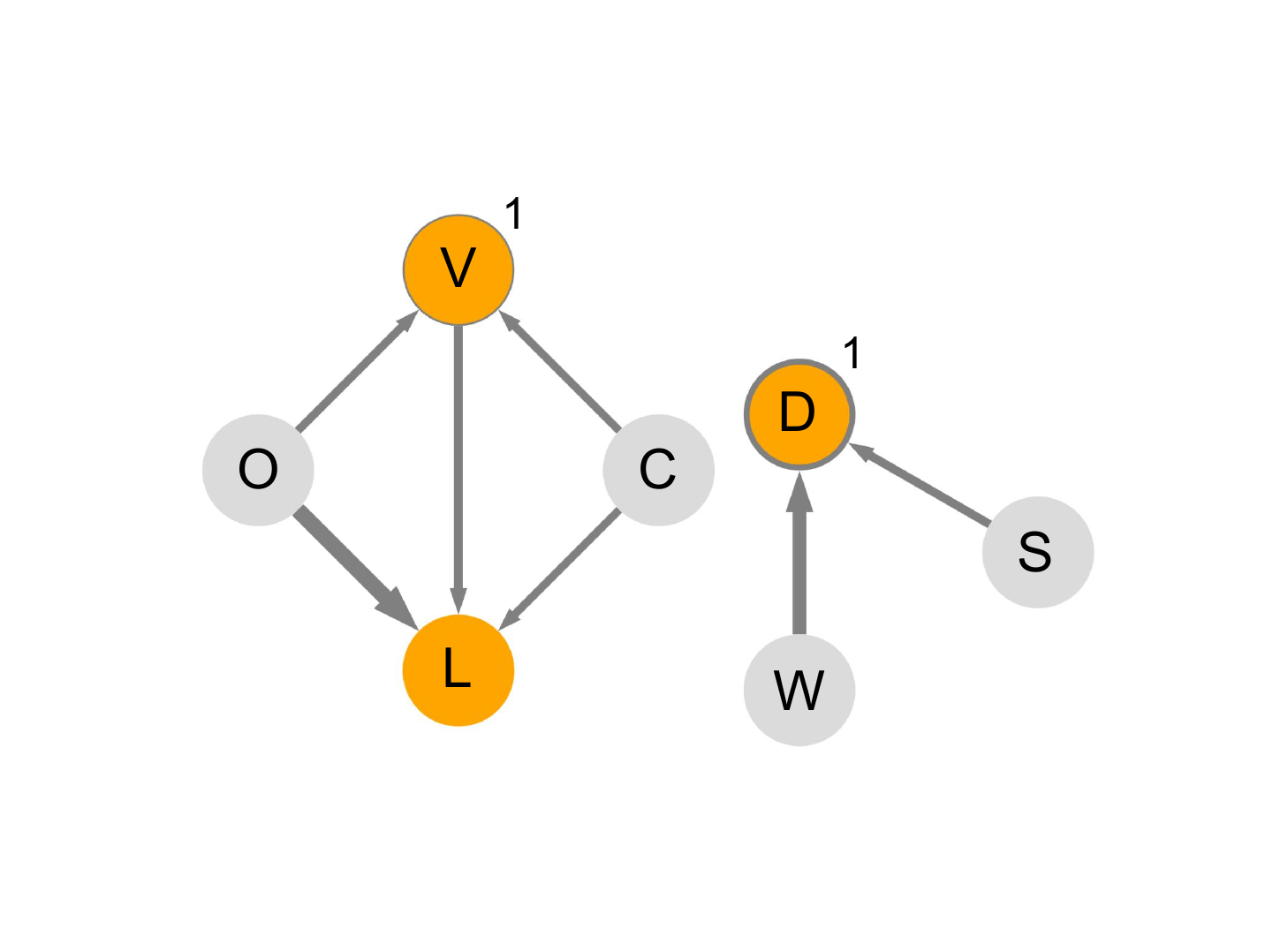}
\caption{Causal model retrieved from the full scenario. Context variables are shown in grey, while system variables are highlighted in orange. The thickness of the arrows and node borders represents the strength of cross- and auto-causal dependencies, respectively, while the numbers on each node and link indicate the time lag of the dependency. To improve readability, arrows without a number represent a contemporaneous effect (a time lag of 0).}
\label{fig:exp-cm}
\end{figure}

The reconstructed causal model, shown in Fig.~\ref{fig:exp-cm}, is consistent with the expected causal relationships that describe the full scenario explained in Sec.~\ref{sec:simulator-scenario} and illustrated in Fig.~\ref{fig:simulator-hypothetical}. As anticipated, the causal model consists of two sub-graphs: one modelling battery dynamics (left part of the figure) and another capturing people density patterns (right part of the figure).
The causal discovery algorithm identified also an additional temporal dependency for the people density ${D_{t-1} \rightarrow D_t}$ (i.e., an autocorrelation). Although not originally planned, this result is plausible, since congestion levels typically persist over time due to gradual crowd movements.
The causal model is then combined with the discretised data to learn the parameters necessary for the inference engine.
As already specified in Sec.~\ref{sec:causal-inference-inference}, these steps of causal discovery and parameters learning are performed only once and are not part of the following decision-making process.

In terms of computational cost, it is well-established that causal discovery scales with the number of variables~\citep{runge_causal_2018,castri2023enhancing,castri2024candoit}. For our specific model, which has 7 variables, the J-PCMCI+ algorithm took 4 minutes and 12 seconds to run on the hardware specified in Sec.~5.

\subsection{Decision-Making Evaluation Setup}\label{sec:exp-evaluation}
As introduced in Sec.~\ref{sec:exp}, our evaluation compares four approaches. The setup for these is as follows. The causal model, shown in Fig.~\ref{fig:exp-cm}, is integrated into the causal inference pipeline of our robot decision-making framework (Fig.~\ref{fig:causal-inference}). As explained in Sec.~\ref{sec:causal-inference-inference} and~\ref{sec:simulator-causal}, this pipeline estimates key quantities that guide path selection: battery consumption ($\hat{L}$) using Eq.~(\ref{eq:causal-inference-delta-rb-exp}) and people density ($\hat{D}$) using Eq.~(\ref{eq:causal-inference-pd-exp}). These estimates are not simply reactive predictions: Eq.~(\ref{eq:causal-inference-delta-rb-exp}) estimates battery consumption adjusting for the confounder $O$, while Eq.~(\ref{eq:causal-inference-pd-exp}) quantifies people density based on time and location.

As a baseline, we benchmark on a common shortest-path heuristic for robot navigation~\citep{nearchou1998path,4339335,tan2021comprehensive}. This approach does not use the causal estimates and is not permitted to refuse tasks. 
This choice of baseline, while simple, is a standard practice in robotics research~\citep{mombelli2022searching,yang2020novel,apoorva2018motion,ng2020adaptive,zhang2024agv} and reflects the default behaviour of widely-used robotics frameworks like ROS, making the comparison relevant to real-world applications.

Our three causal-aware approaches use the estimates ($\hat{L}$ and $\hat{D}$) to enhance this baseline:
\begin{itemize}
    \item \textbf{Causal Routing}: uses the full causal heuristic (Eq.~\ref{eq:causal-inference-heuristic}) but is forced to accept all tasks (i.e., the refusal mechanism in Eq.~(\ref{eq:total-battery-consumption}) is disabled).
    \item \textbf{Refusal-Only}: uses the causal task refusal mechanism (Eq.~\ref{eq:total-battery-consumption}) to abort tasks but relies on the simple shortest-path heuristic for routing.
    \item \textbf{Full Causal Framework}: employs both the causal heuristic (Eq.~\ref{eq:causal-inference-heuristic}) and the causal task refusal mechanism (Eq.~\ref{eq:total-battery-consumption}).
\end{itemize}

An experiment is conducted for each time-slot~$S$. Workers move according to the waypoint probability distribution defined for~$S$, as shown in Fig.~\ref{fig:exp-scenario}. Similarly, the robot performs a predetermined number of tasks, always following the plan outlined in the figure. Regardless of the $S$ duration, each experiment concludes when the robot completes all assigned tasks for that time-slot, so that baseline and causal approaches are compared on the same number of tasks rather than how many tasks the robot manages to complete within $S$. To eliminate potential biases, each experiment begins with the robot’s battery fully charged. Additionally, we predefine the workers' goal sequences to match the scenarios in Fig.~\ref{fig:exp-scenario}, ensuring that the only difference between the two approaches is the heuristic used for path planning.
We set a $45$ seconds deadline for task completion: if the robot fails to complete a task within this time limit, it is considered a failure, and the robot is moved to the next task's starting position. This prevents the robot from being indefinitely stuck due to nearby human presence, while also ensuring that each task starts from the same location for both approaches.

From $S 1$ to $S 10$, the number of tasks is set to 200 in order to collect enough interaction data for assessing the benefits of $D$ and battery consumption estimations, as well as evaluating the effect of causal decision-making process.
%
%
In $S 11$, when there are no workers, the robot navigates through all the waypoints in the environment, simulating a simple cleaning routine during non-working hours. Each connection between waypoints is considered as a task that the robot must complete. The minimum number of arcs connecting all the waypoints was determined using the Traveling Salesman Problem~\citep{gutin2006traveling}, resulting in 90 connections, corresponding to 90 tasks.
%

Following a comprehensive sensitivity analysis (presented in~\ref{app:sensitivity}), we selected the heuristic weights: ${\lambda_\delta = 1}$, ${\lambda_{D} = 10}$, and ${\lambda_{L} = 5}$. These values were chosen to ensure the robot favours safer and less crowded paths, prioritising congestion avoidance over minimising the total travelled distance.
If the battery drops below a critical threshold (i.e., $20\%$ of the total charge) during the execution of a task, this is considered failed, and the battery is reset to~$100\%$. If the battery falls below the critical threshold after a task has completed, it is still reset but considered a success. 
Note that for the causal approach, if our decision making framework does not return any valid path to a given goal that satisfies the critical battery constraint, the battery level is set again to $100\%$, as if the robot had been at the charging station.

\subsection{Metrics}\label{sec:exp-metrics}
Besides principles for social navigation, \citet{francis2023principles} specifies metrics for their assessment, some of which are used to evaluate the {\bf efficiency} of our approach:
\begin{itemize}
    \item \textit{Success Rate:} the proportion of successfully completed tasks out of their total number.
    \item \textit{Failure Rate:} the proportion of failed tasks out of their total number, further categorised into failures due to battery depletion and to congestion.
    \item \textit{Task Time:} for successful tasks, this metric measures both the time during which the robot is actively moving toward its goal and the time during which it is stalled (i.e., null velocity). For failed tasks, it measures the time wasted without reaching the goal.
    \item \textit{Travelled Distance:} for successful tasks, this metric quantifies the total distance travelled by the robot, split into planned path length (as computed by the A\textsuperscript{*} algorithm) and extra distance incurred due to unexpected deviations. For failed tasks, it measures the travel distance wasted without reaching the goal.
    \item \textit{Battery Usage:} battery consumed during completed tasks or incomplete attempts.
\end{itemize}

To evaluate the improved {\bf safety} in human-robot shared environments, we considered instead the following metrics:
\begin{itemize}
    \item \textit{Collision Count:} measures the number of robot-human collisions, considered as such when the distance between the robot and a person falls below a predefined safety threshold~\citep{mavrogiannis2023core}. Specifically, this metric counts instances where a worker enters the robot’s circumscribed region\footnote{\url{http://wiki.ros.org/costmap_2d}\label{foot:costmap}}.

    \item \textit{Space Compliance:} evaluates the robot’s adherence to human spatial preferences based on Hall’s proxemic zones~\citep{hall1968proxemics}.
\end{itemize}

To assess the significance of the differences between our causal approach and the baseline, we apply opportune statistical tests based on the nature of each metric~\citep{rabe2003handbook}. In particular, for the Success and Failure Rate, we use a Chi-Square test, which is appropriate for categorical outcomes (success vs. failure). For the Collision Count, we employ a Negative Binomial test, suitable for count data that exhibits overdispersion (i.e., variance greater than the mean), as observed in our case. For all the other continuous or ordinal metrics---Task Time, Travelled Distance, Battery Usage, and Space Compliance---we use the Mann-Whitney U test, a non-parametric alternative to the t-test, since it does not assume normally distributed data.

\subsection{Results}\label{sec:exp-results}
\begin{figure}[!t]
    \centering
    \begin{subfigure}{0.49\textwidth}
        \includegraphics[trim={0cm 0cm 0cm 0cm}, clip, width=\columnwidth]{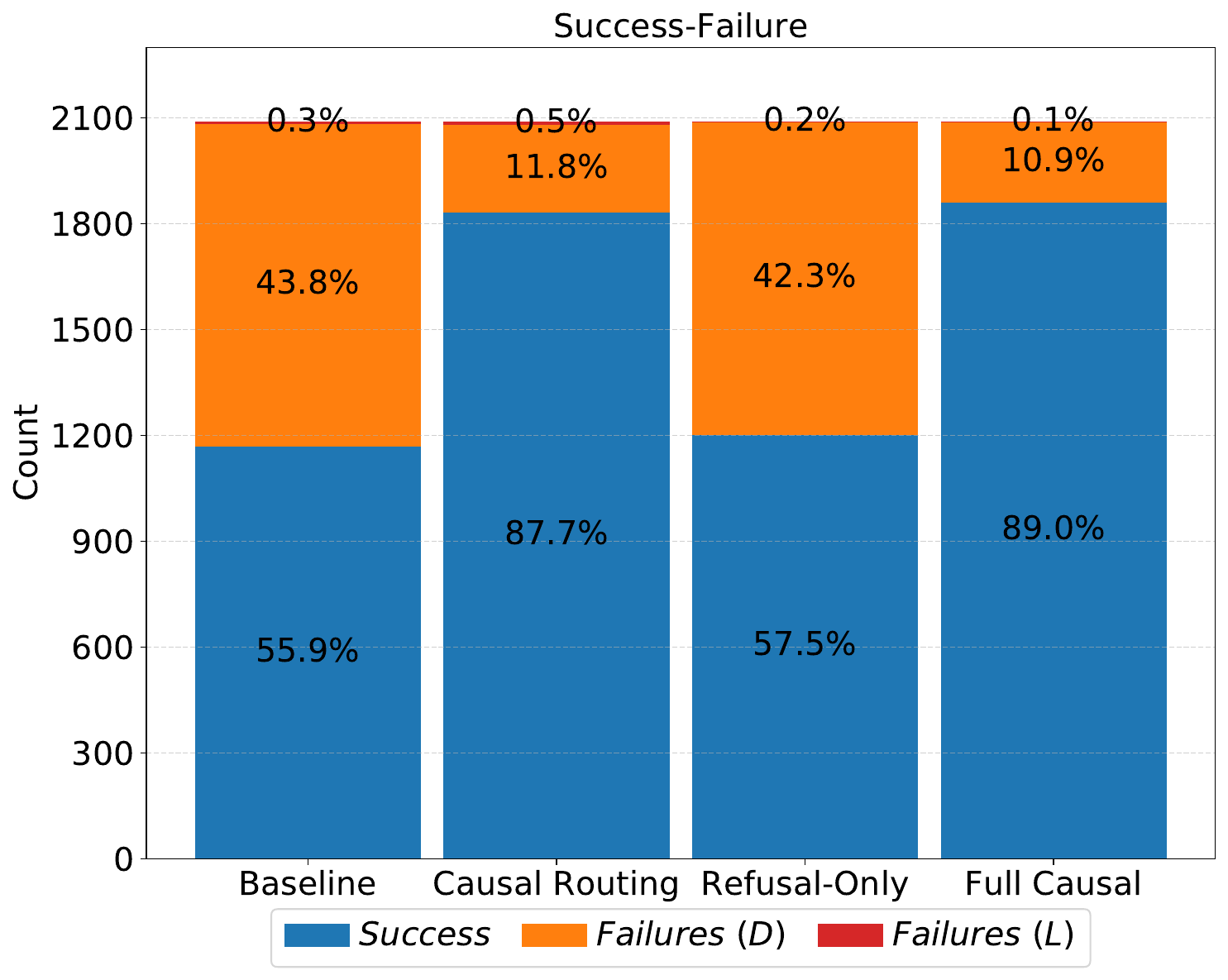}
        \caption{}\label{fig:res-efficiency-sf}
    \end{subfigure}
    \begin{subfigure}{0.495\textwidth}
        \includegraphics[trim={0cm 0cm 0cm 0cm}, clip, width=\columnwidth]{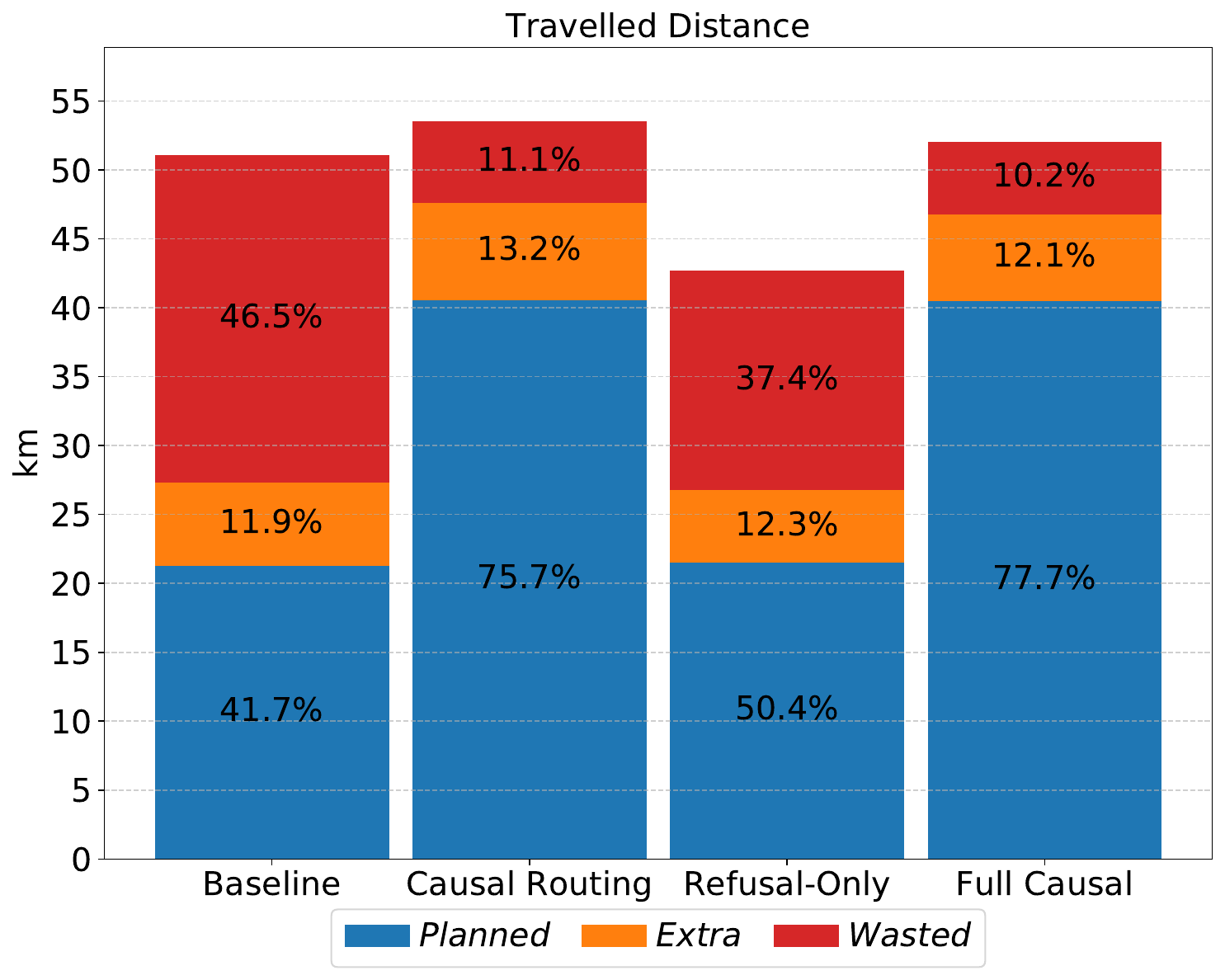}
        \caption{}\label{fig:res-efficiency-dist}
    \end{subfigure}\\[0.1cm]
    \begin{subfigure}{0.49\textwidth}
        \includegraphics[trim={0cm 0cm 0cm 0cm}, clip, width=\columnwidth]{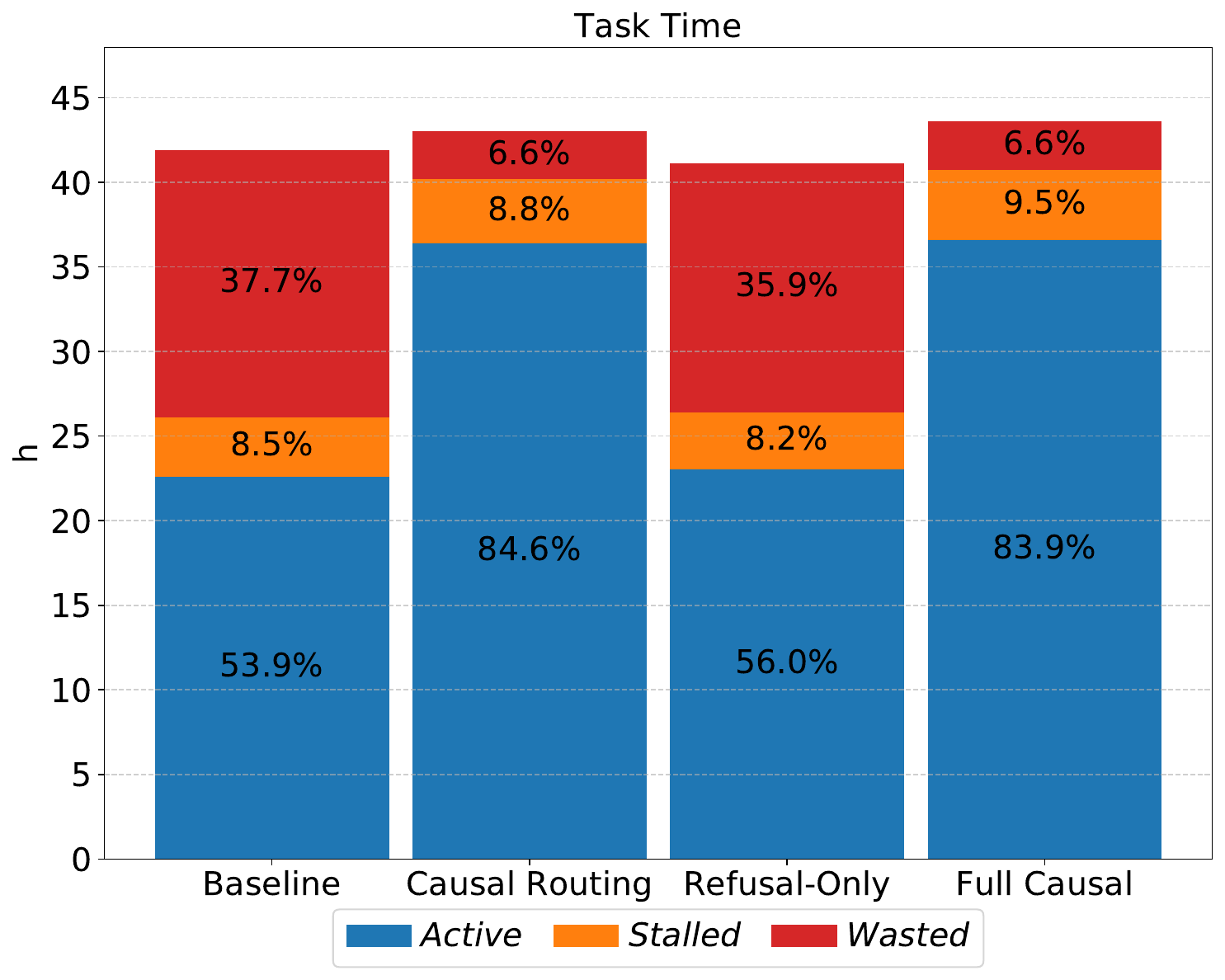}
        \caption{}\label{fig:res-efficiency-time}
    \end{subfigure}
    \begin{subfigure}{0.49\textwidth}
        \includegraphics[trim={0cm 0cm 0cm 0cm}, clip, width=\columnwidth]{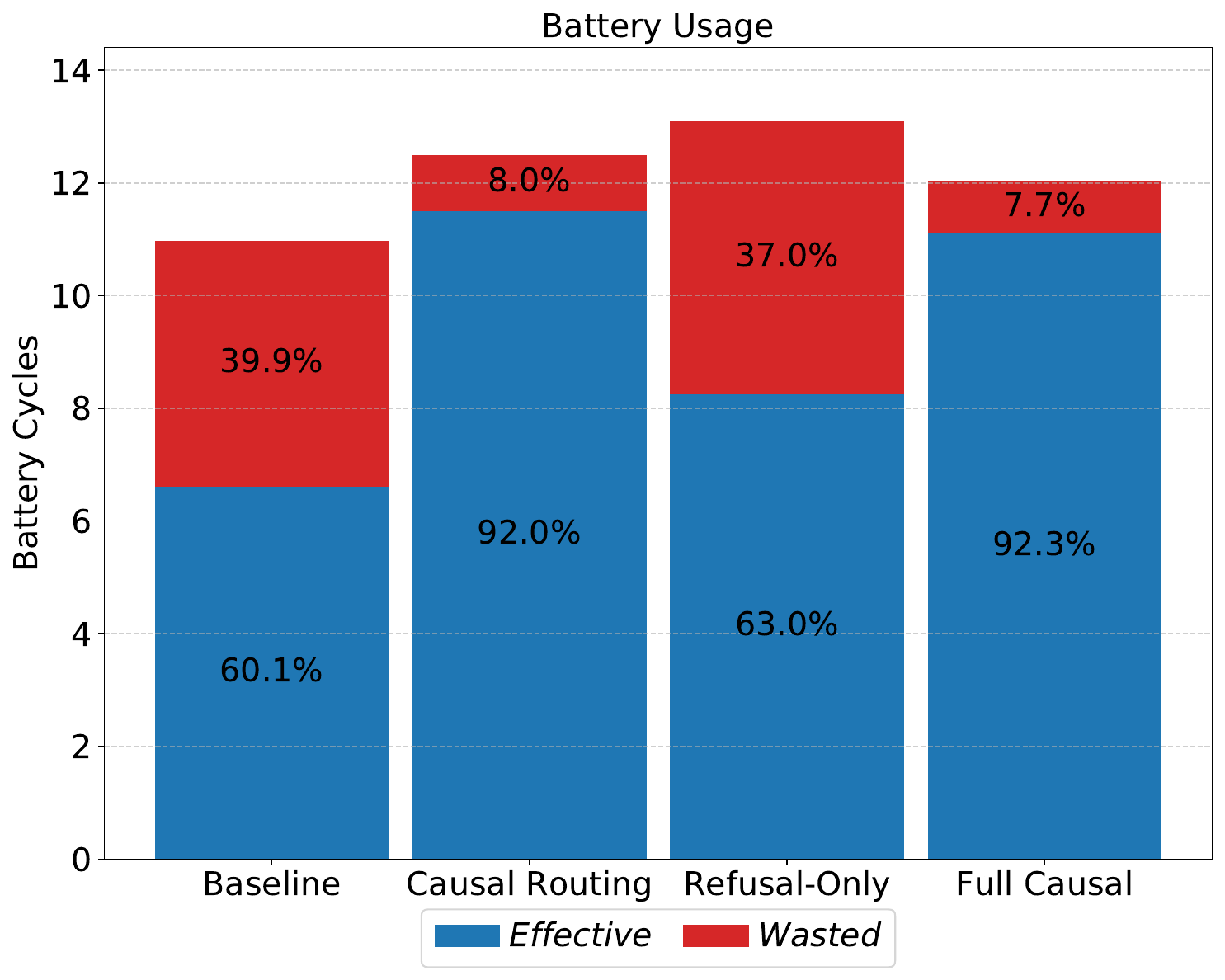}
        \caption{}\label{fig:res-efficiency-battery}
    \end{subfigure}
    \caption{Summary of overall efficiency metrics across all time-slots. 
    (a) Task outcomes: (blue) successful tasks, (orange) failures due to high people density $D$, and (red) failures caused by excessive battery consumption $L$. 
    (b) Travelled distance breakdown: (blue) planned distance (expected travel), (orange) extra distance (deviations from the expected), and (red) wasted distance linked to failed tasks. 
    (c) Task completion time distribution: (blue) active time (robot moving during successful tasks), (orange) stalled time (robot idle in successful tasks), and (red) wasted time (time spent in failed tasks). 
    (d) Battery usage analysis: (blue) effective battery consumption in successful tasks and (red) wasted battery in failed tasks.
    %
    }
    \label{fig:res-efficiency}
\end{figure}

We present the results of our experiments, highlighting the benefits of our approach in terms of {\em efficiency} and {\em safety}.  
\paragraph{\textbf{Efficiency}}
As outlined in Sec.~\ref{sec:exp-metrics}, the efficiency results shown in Fig.~\ref{fig:res-efficiency} include the following metrics: (\ref{fig:res-efficiency-sf}) task success and failure counts, (\ref{fig:res-efficiency-dist}) travelled distance, (\ref{fig:res-efficiency-time}) task completion time, and (\ref{fig:res-efficiency-battery}) battery consumption. Each graph directly compares the four approaches from our ablation study (Sec.~\ref{sec:exp-evaluation}): the \textbf{Baseline}, \textbf{Causal Routing}, \textbf{Refusal-Only}, and the \textbf{Full Causal} framework. Each approach is represented by a stacked bar, which is divided into multiple blocks representing different categories within that metric. The percentage shown in each block is relative to the total height of its bar, providing a clear visual representation of the proportion of each category.
\begin{table}[b]
\centering
\caption{Baseline vs Causal runtime performance.}
\label{tab:runtime_performance}
\begin{tabular}{@{}lcc@{}}
\toprule
\textbf{Hardware} & \multicolumn{2}{c}{
                        \begin{tabular}{@{}l@{}}
                         CPU: 12th Gen Intel Core i9-12900F \\
                         RAM: 32GB \\
                         GPU: RTX 4070 \\
                         OS: Ubuntu 20.04~LTS \\
                         ROS: Noetic
                        \end{tabular}} \\
\midrule
\textbf{Metric} & \textbf{Baseline} & \textbf{Causal} \\
\midrule
Avg. Query Time (s) & 0 (N/A) & 0.327 \\
Avg. Candidate Routes (count) & 18.20 & 25.46 \\
\bottomrule
\end{tabular}
\end{table}
Starting with number of successful and failed tasks~(Fig.~\ref{fig:res-efficiency-sf}), we distinguish the latter into failures due to the robot getting stuck because of people congestion~(orange) and failures caused by the battery level dropping below the critical threshold~(red). 

The travelled distance~(Fig.~\ref{fig:res-efficiency-dist}) is also divided into three components: the total planned path length across all time-slots~(blue), representing the total length of the paths determined by the heuristic; extra travelled distance~(orange), which quantifies deviations from the planned paths, computed as the difference between the actual total travelled distance and the total planned distance; and wasted distance~(red), corresponding to the distance covered during tasks that ultimately failed. The planned and extra travelled distances are only measured for successfully completed tasks. This metric is shown in kilometres~(km).  

The task time~(Fig.~\ref{fig:res-efficiency-time}) is divided into three components: active time~(blue), representing the total duration in which the robot is actively moving toward its goal (i.e., performing tasks); stalled time (orange), accounting for periods when the robot stopped due to congestion or obstacle avoidance; and wasted time (red), representing the total time spent on tasks that ultimately resulted in a failure. Note that active and stalled times are only measured for successfully completed tasks. This metric is measured in hours (h).  

The battery usage (Fig.~\ref{fig:res-efficiency-battery}) is divided into two categories: effective battery usage (blue), representing the battery consumed during successfully completed tasks, and wasted battery usage (red), which quantifies the battery consumed in tasks that ultimately resulted in a failure. Battery usage is measured in battery cycles, where one cycle corresponds to consuming $100\%$ of the battery.

In addition to task-level efficiency, we measured the computational cost of the causal query loop, as reported in Table~\ref{tab:runtime_performance}. The table summarises the hardware configuration used for the experiments and compares the baseline and causal approaches. In particular, it highlights the runtime performance of the two methods, showing the average query time across all experiments and the average number of candidate routes evaluated by the heuristic functions.

\begin{figure}[t]
    \centering
    \begin{subfigure}{0.495\textwidth}
        \includegraphics[trim={0cm 0cm 0cm 0cm}, clip, width=\columnwidth]{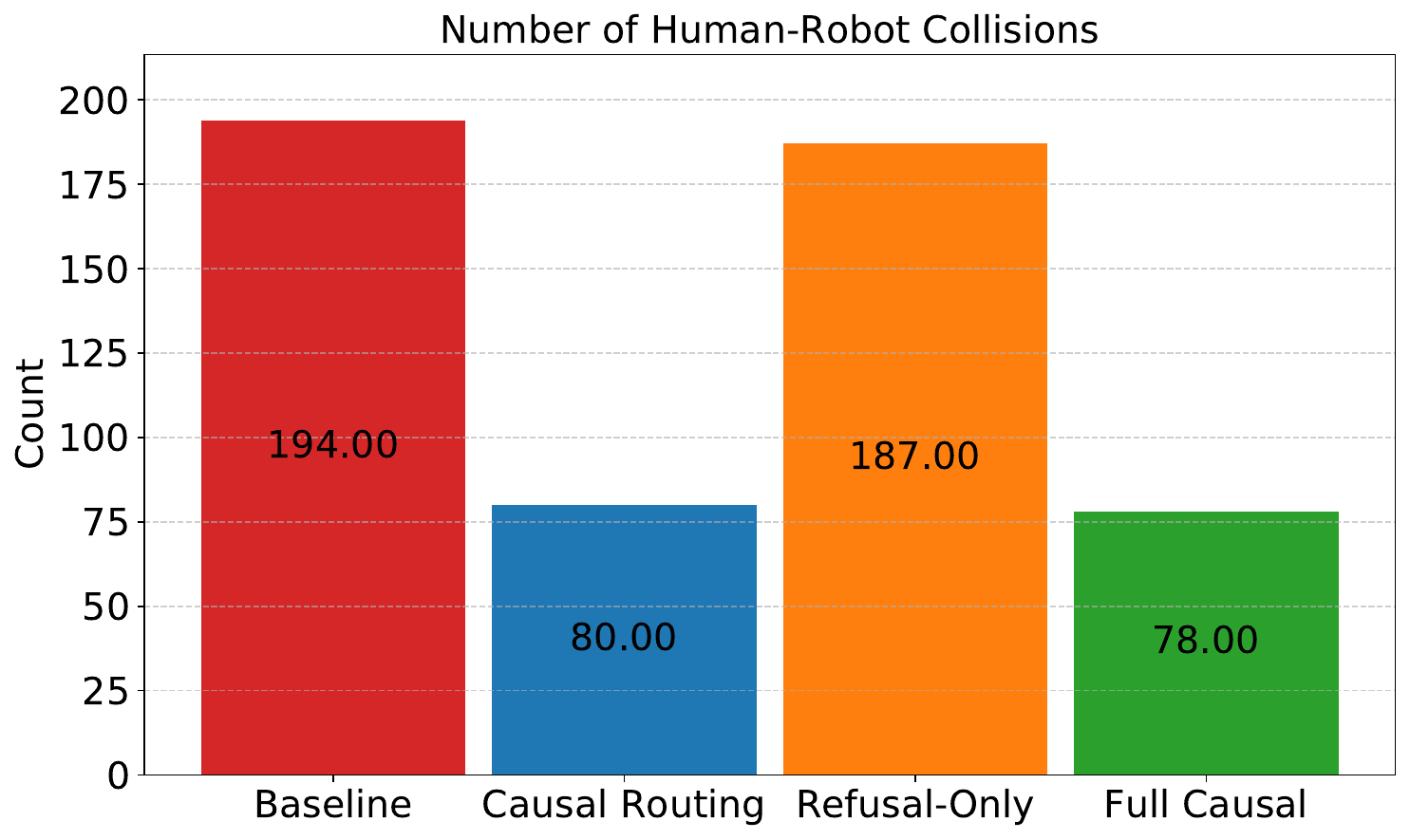}
        \caption{}\label{fig:res-safety-dang}
    \end{subfigure}
    \begin{subfigure}{0.495\textwidth}
        \includegraphics[trim={1cm 0cm 0cm 0cm}, clip, width=\columnwidth]{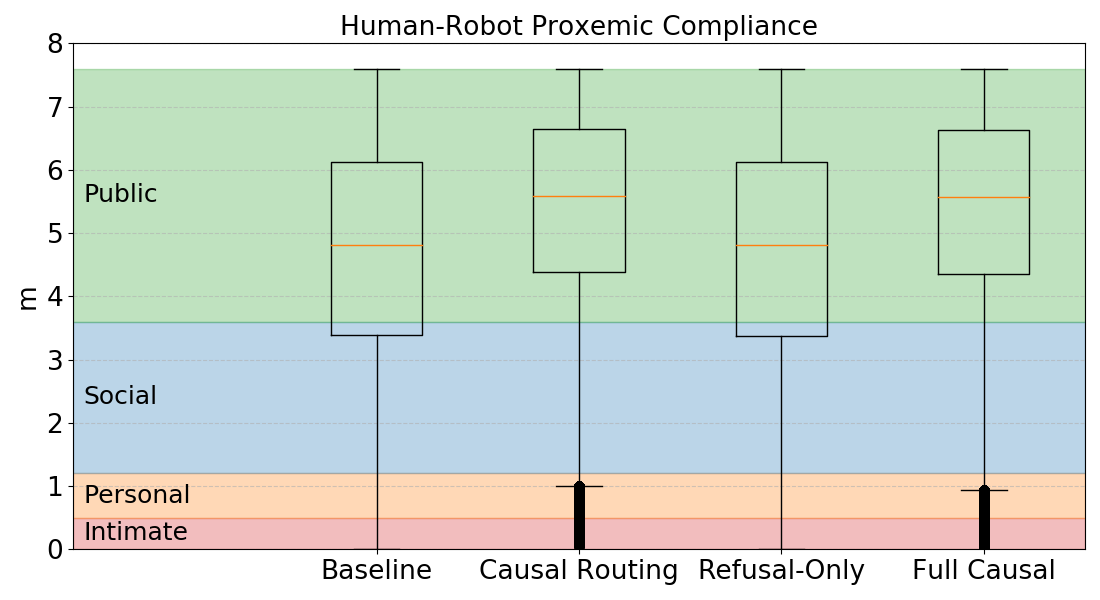}
        \caption{}\label{fig:res-safety-proxemics}
    \end{subfigure}
    \caption{Summary of overall safety results across all time-slots. (a)~Number of human-robot collisions of the various approaches: Baseline~(red), Causal Routing~(blue), Refusal-Only~(orange), and Full Causal~(green). (b)~Human-robot proxemic compliance with proxemic zone bands: (green)~public, (blue)~social, (orange)~personal, (red)~intimate. 
    %
    }
    \label{fig:res-safety}
\end{figure}

\paragraph{\textbf{Safety}} 
Sec.~\ref{sec:exp-metrics} defines the metrics to evaluate the safety of HRSI, shown in Fig.~\ref{fig:res-safety}, and including (\ref{fig:res-safety-dang})~number of collisions and (\ref{fig:res-safety-proxemics})~human-robot proxemic compliance.
The number of collisions quantifies instances where robot and human are within contact distance, which is TIAGo’s base radius\footref{foot:TIAGo}. The graph compares the following approaches: Baseline~(red), Causal Routing~(blue), Refusal-Only~(orange), and Full Causal~(green).
Fig.~\ref{fig:res-safety-proxemics} shows a box plot of the distribution of human-robot distances across all time-slots. The background highlights different proxemic zones~\citep{hall1968proxemics}: public~(${\leq7.6\text{m}}$, green), social~(${\leq3.6\text{m}}$, blue), personal~(${\leq1.2\text{m}}$, yellow), and intimate~(${\leq0.5\text{m}}$, red).

\subsection{Discussions}\label{sec:exp-discussions}
The results introduced in Sec.~\ref{sec:exp-results} are discussed here in detail to highlight the benefits of our proposed approach.  

\begin{figure}
    \centering

    \begin{subfigure}{0.343\textwidth}
        \includegraphics[trim={1cm 0cm 1cm 10cm}, clip, width=\columnwidth]{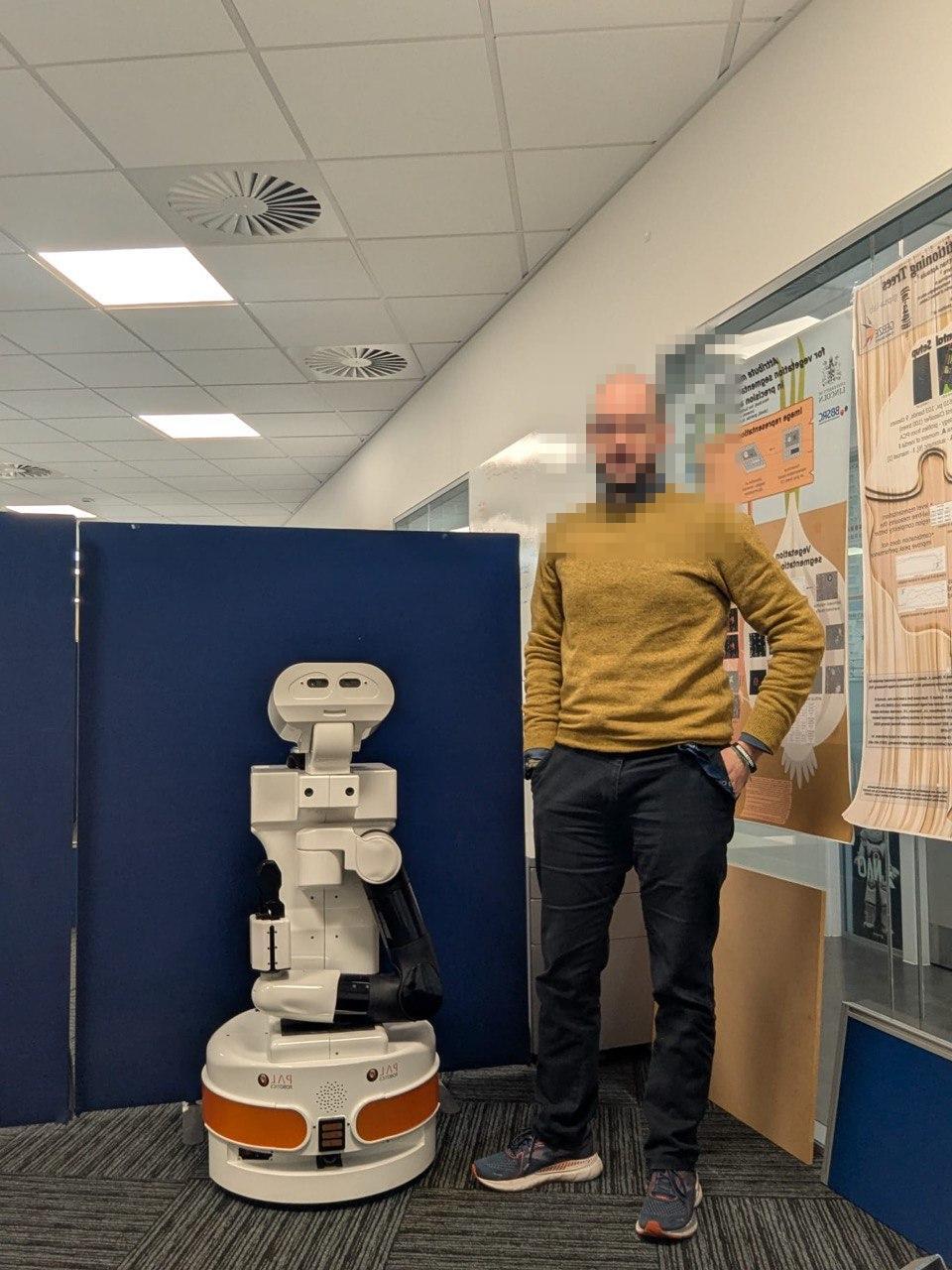}
        \caption{}\label{fig:exp-examples-realtiago}
    \end{subfigure}\hfill
    \begin{subfigure}{0.64\textwidth}
        \includegraphics[trim={12cm 10cm 14cm 3.4cm}, clip, width=\columnwidth]{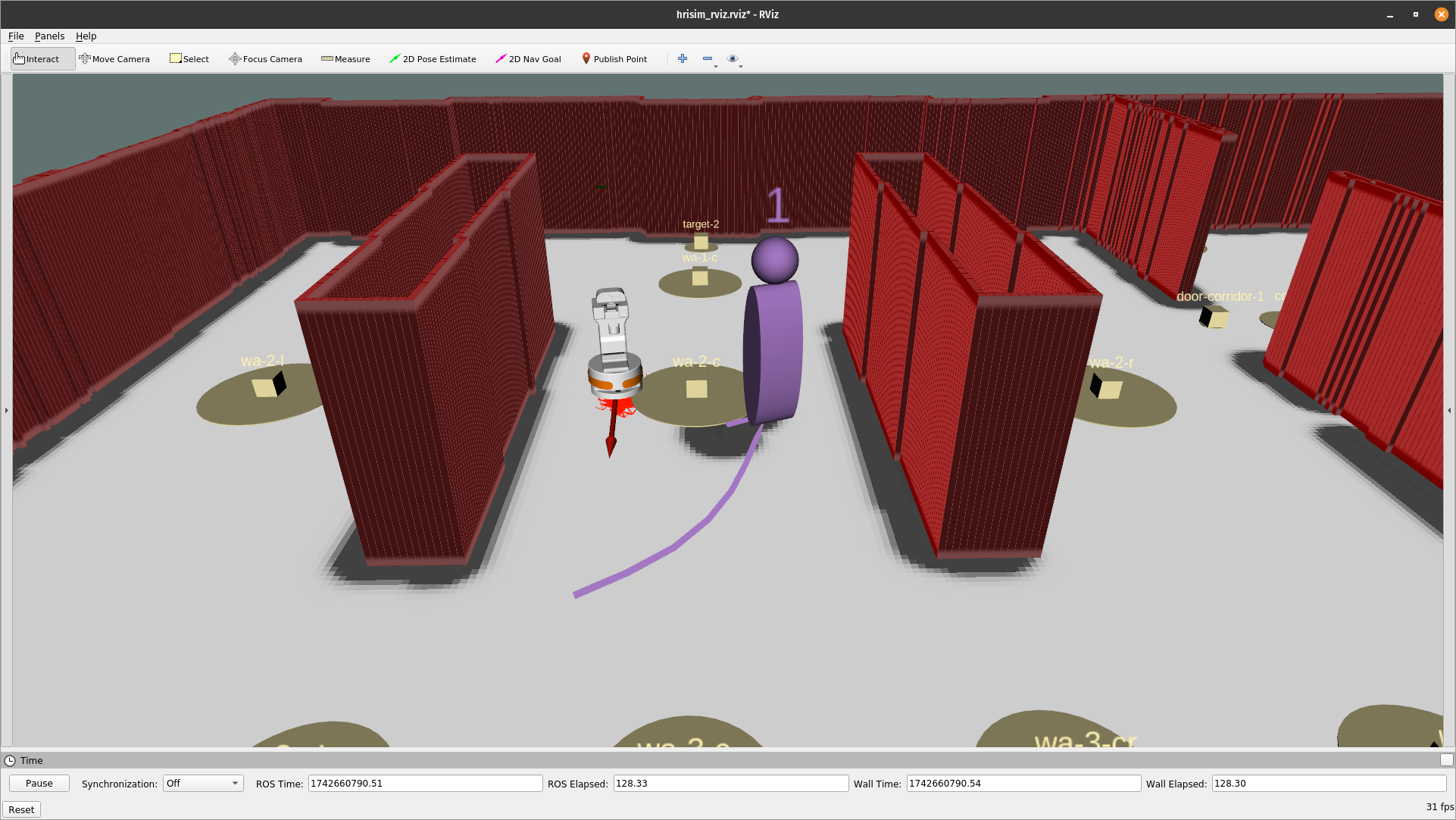}
        \caption{}\label{fig:exp-examples-simtiago}
    \end{subfigure}\\
    \begin{subfigure}{0.49\textwidth}
        \includegraphics[trim={10cm 3.6cm 10cm 3.5cm}, clip, width=\columnwidth]{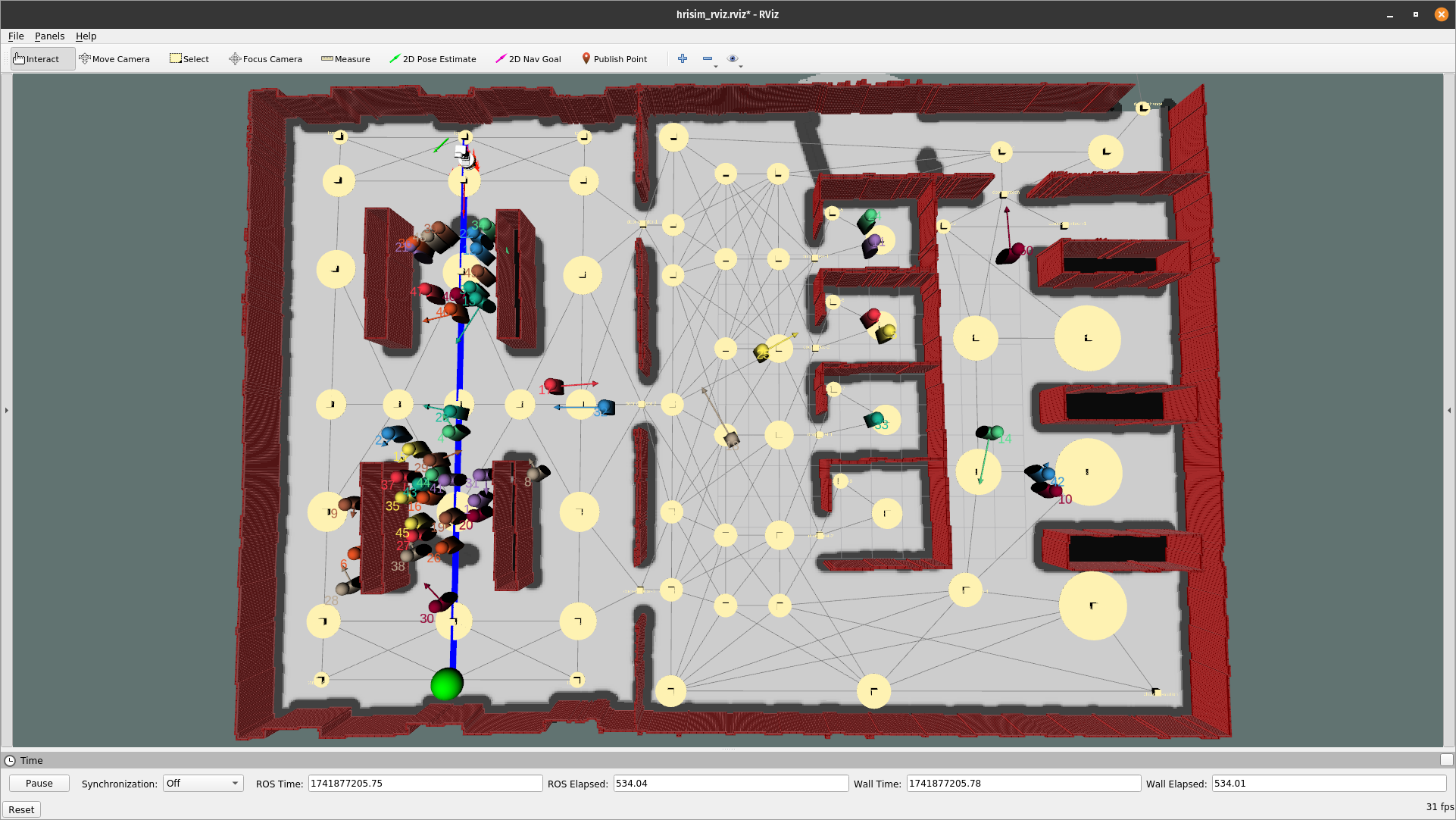}
        \caption{}\label{fig:exp-examples-bases2top}
    \end{subfigure}\hfill
    \begin{subfigure}{0.49\textwidth}
        \includegraphics[trim={10cm 3.4cm 9.5cm 3.5cm}, clip, width=\columnwidth]{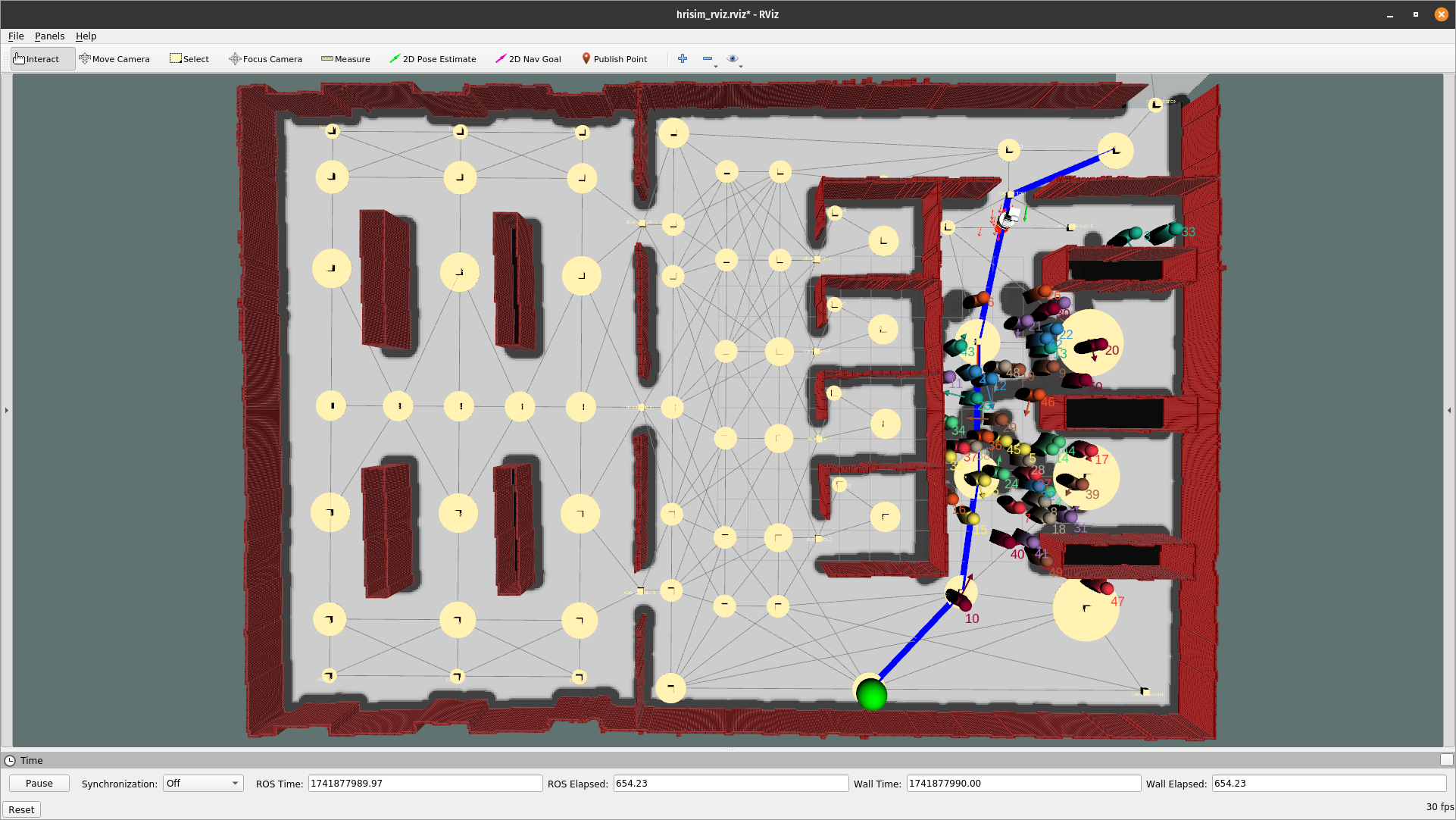}
        \caption{}\label{fig:exp-examples-bases6top}
    \end{subfigure}\\
    \begin{subfigure}{0.49\textwidth}
        \includegraphics[trim={10cm 3.6cm 10cm 3.5cm}, clip, width=\columnwidth]{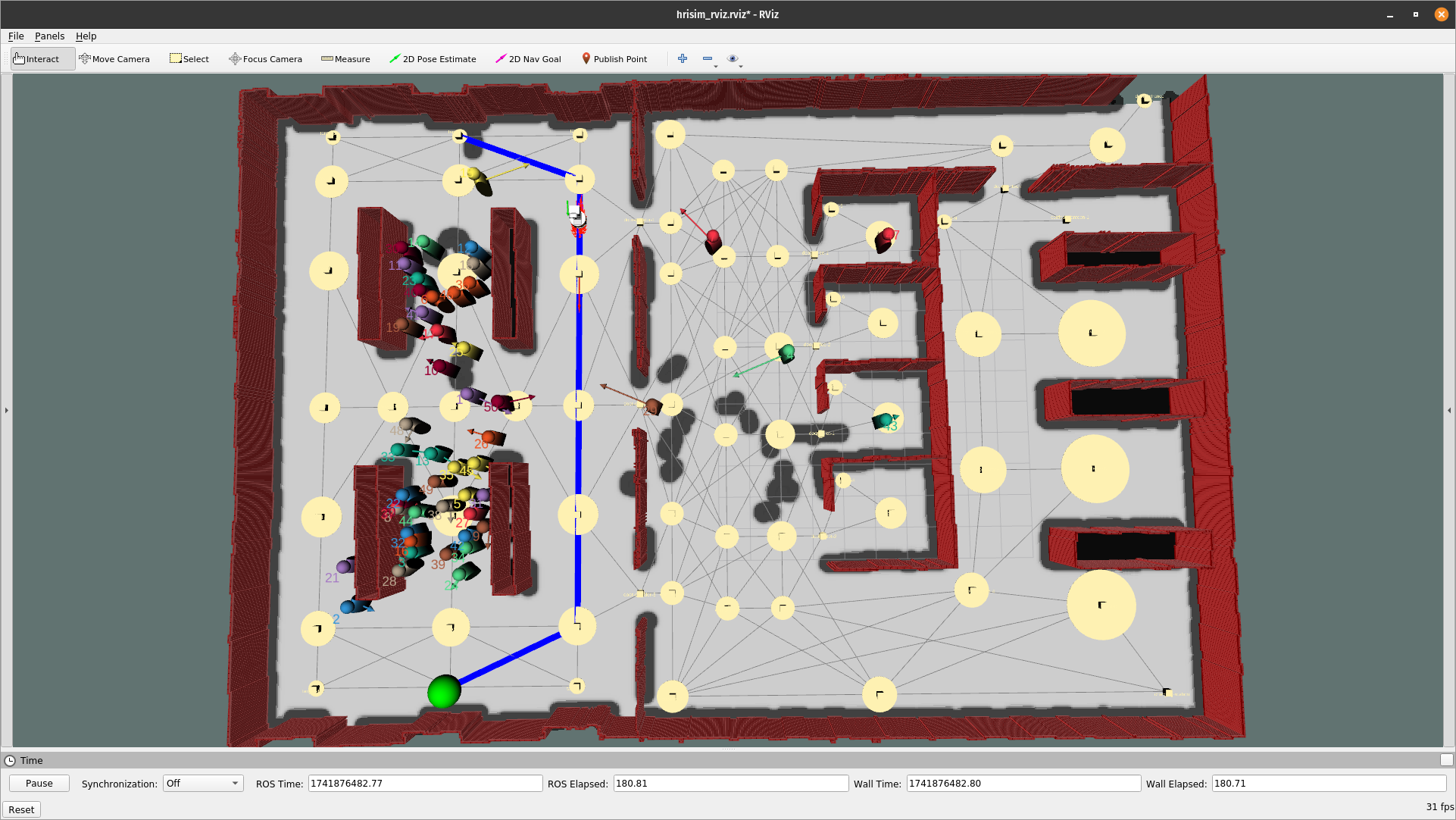}
        \caption{}\label{fig:exp-examples-causals2top}
    \end{subfigure}\hfill
    \begin{subfigure}{0.49\textwidth}
        \includegraphics[trim={10.2cm 3.6cm 10cm 3.5cm}, clip, width=\columnwidth]{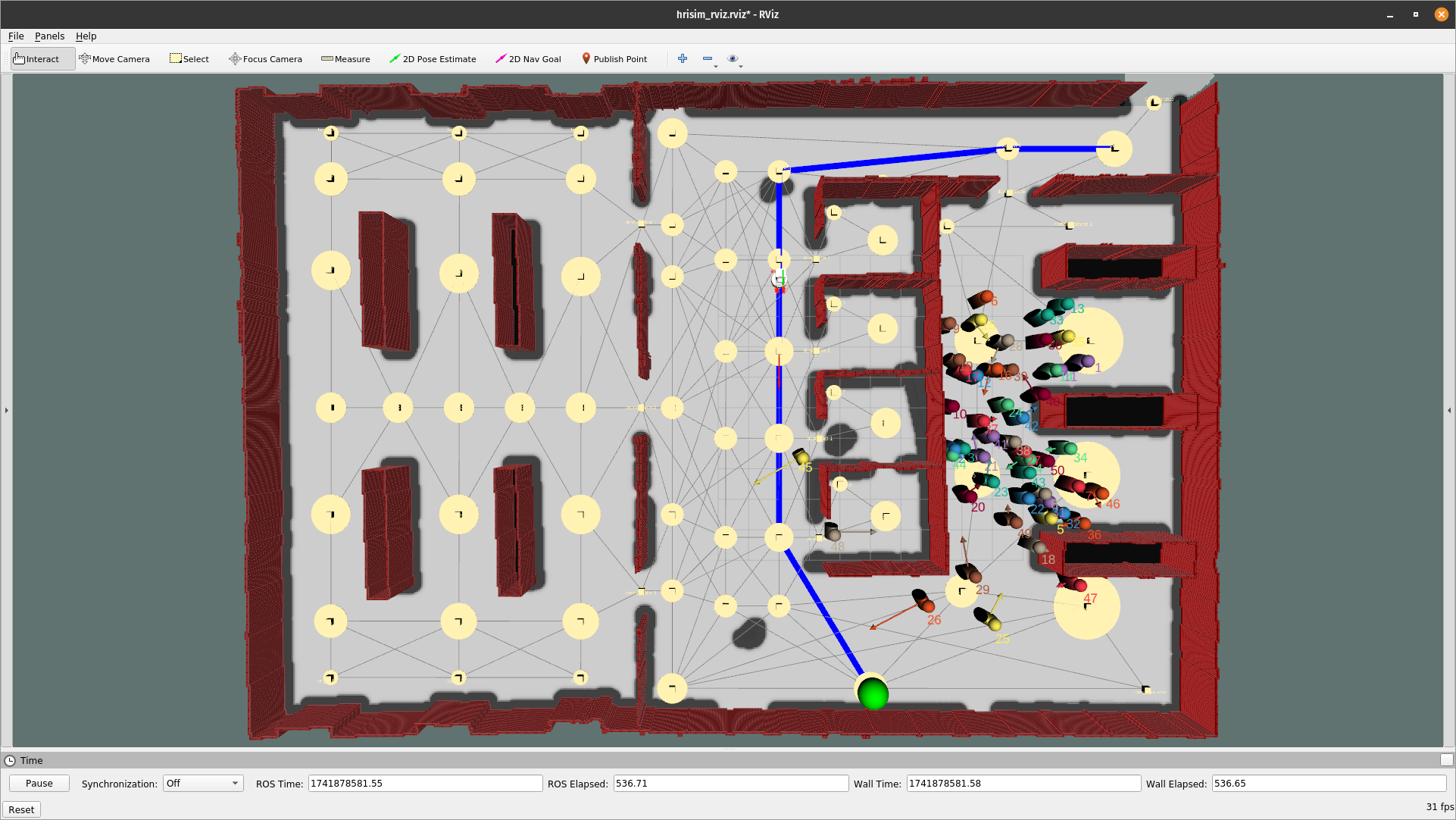}
        \caption{}\label{fig:exp-examples-causals6top}
    \end{subfigure}
    \caption{
    (a) TIAGo robot with a person in a real-world scenario. (b) TIAGo robot with a simulated agent in PeopleFlow.
    (c, d, e and f) Visualisations of the TIAGo robot executing its assigned tasks in two scenarios: (c, e) $S 2$ working time and (d, f) $S 6$ lunchtime. The robot's goal is represented by a green sphere, while the planned path is highlighted in blue. Congested areas align with Fig.~\ref{fig:exp-scenario} in both scenarios. 
    %
    %
    (c, d) Path generated by the shortest-strategy heuristic in the baseline approach.
    (e, f) Path generated by our approach. The bottom-line graphs illustrate the effectiveness of our method, which prioritises safety over minimising travel distance.}
    \label{fig:exp-examples}
\end{figure}

\paragraph{\textbf{Efficiency}}
Fig.~\ref{fig:res-efficiency-sf} now clearly illustrates the benefits of our framework and the distinct contributions of its components. The Full Causal framework achieves the highest success rate at $89.0\%$ ($1860$ out of $2090$ tasks), a substantial improvement compared to the baseline's $55.9\%$. The ablation study shows the source of this gain is due to the use of exploiting the causal inference engine for the path planning (Eq.~\ref{eq:causal-inference-heuristic}). Indeed, the Causal Routing approach is clearly the primary driver of improved performance. By itself, it boosts the success rate to $87.7\%$, almost entirely by reducing congestion failures (Failures D) from $43.8\%$ (Baseline) to $11.8\%$. In contrast, the Refusal-Only approach, which still uses the shortest-path router, provides almost no benefit ($57.5\%$ success), proving that simply refusing tasks is not an effective strategy when the routing itself is poor. However, Causal Routing is still vulnerable to battery-related failures, which occur in $0.5\%$ of the tasks, while the Refusal-Only approach, which has the refusal mechanism, suffers only $0.2\%$ battery failures. This shows a clear synergistic benefit in our Full Causal framework: by combining smart routing with the refusal mechanism, it catches the edge cases, reducing battery failures to just $0.1\%$ and achieving the highest overall success rate. This demonstrates that both estimations of people density~($D$) and battery consumption~($L$) are necessary, with the causal heuristic eliminating the majority of congestion failures and the refusal mechanism providing a critical safeguard against battery-related failures.
An example of the experiments in Fig.~\ref{fig:exp-examples} shows a typical scenarios where TIAGo, using our approach, successfully avoids congested areas and completes its task. In contrast, the baseline approach, which prioritises the shortest path---even when it is the busiest---leads the robot into congestion, ultimately causing task failure.

%

The overall distance travelled by the robot is illustrated in Fig.~\ref{fig:res-efficiency-dist}.
%
The impact of these path selection strategies is evident in the wasted distance (red blocks), which represents the distance travelled in tasks that ultimately failed. 
The Baseline approach, which relies on the shortest-path strategy, results in $46.5\%$ of its total travelled distance being wasted. The ablation study clearly shows that the Causal Routing heuristic is the primary solution to this: by itself, it drastically reduces 
wasted distance to just $11.1\%$. In contrast, the Refusal-Only approach still wastes $37.4\%$ of its distance, once again proving that smart refusal is ineffective without smart routing. The Full Causal framework combines both, achieving the lowest wasted distance of all at $10.2\%$, demonstrating a significant improvement in motion efficiency.

These findings are consistent with the task time results in Fig.~\ref{fig:res-efficiency-time}. 
Specifically, the wasted time (red blocks)---the time spent on tasks that ultimately failed---is greatly reduced. 
The Baseline approach wastes $37.7\%$ of its total time on failed tasks. The ablation study again shows the Causal Routing heuristic is the primary driver of performance, by itself drastically reducing wasted time to just $6.6\%$. The Refusal-Only approach, in contrast, shows almost no improvement, remaining at $35.9\%$ wasted time. The Full Causal framework ($6.6\%$ wasted) performs identically to the Causal Routing approach, confirming that the smart heuristic is the main source of this efficiency gain.
This reduction in waste directly translates to a massive increase in productive work. The robot actively moves toward its goals (blue block) for $83.8\%$ and $84.6\%$ of the total time in the Full Causal framework and in Causal Routing, respectively, compared to 
only $53.6\%$ for the Baseline and $56\%$ for the Refusal-Only.

Finally, the battery usage in Fig.~\ref{fig:res-efficiency-battery} highlights the framework's impact on energy efficiency. The Baseline approach wastes a significant $39.9\%$ of its total battery consumption on failed tasks. The ablation study clearly shows that the Causal Routing heuristic is the primary solution to this problem. By itself, it reduces wasted battery to just $8\%$. In contrast, the Refusal-Only approach is aligned with the baseline result, wasting $37.0\%$ of its battery. The Full Causal framework achieves the best result, wasting only $7.7\%$ of its battery. This slight difference over Causal Routing ($8.0\%$) is important: it confirms the synergistic advantage of the refusal mechanism. As shown in Fig.~\ref{fig:res-efficiency-sf}, the refusal component catches the few catastrophic battery failures (Failures L) that Causal Routing cannot preemptively avoid, making the Full Causal framework the most robust. This is a key finding: our framework's battery usage is significantly better managed, dedicating $92.3\%$ of its consumption to successful tasks, compared to only $60.1\%$ for the baseline.

Note that, in $S11$ (off-time), both methods behave identically. Specifically, in all approaches, the robot successfully completes all $90$ tasks, travelling comparable distances in the same amount of time and using the same amount of battery. This demonstrates that our approach does not introduce unnecessary path deviations in the absence of workers and congestion.

Statistical significance was confirmed by a Chi-Square test for the success-failure graph and a Mann-Whitney U test for the other metrics, as explained in Sec.~\ref{sec:exp-metrics}.

Beyond task-level efficiency, we also measured the computational cost of the inference pipeline. The average inference time per task was approximately $0.32$~s, as detailed in Table~\ref{tab:runtime_performance}. The table compares two runtime performance metrics: the average query time and the average number of candidate routes per task across all time slots. In the causal case, the number of candidate paths evaluated by the heuristic function is slightly higher compared to the baseline approach. The query time for the baseline is, by definition, $0$~s, whereas for the causal approach it is $0.327$~s. This confirms that the computational overhead introduced by our causal query remains low enough for practical, real-time planning. It is important to note that this time was recorded on hardware that was simultaneously running the inference code and the computationally expensive Gazebo physics simulation. On a real robot's onboard computer, which would only run the inference, this query time would be lower, confirming its suitability for real-time planning. As demonstrated in our scalability analysis (see~\ref{app:scalability}), the query cost scales linearly with the size of the map. This predictable, linear behaviour confirms its suitability for real-time planning applications, as the cost does not explode with map complexity.


%

\paragraph{\textbf{Safety}}
Fig.~\ref{fig:res-safety} presents the safety-related results, where the ablation study highlights the source of our framework's benefits.
Fig.~\ref{fig:res-safety-dang} reveals a substantial improvement in the human-robot collision metric. The Baseline approach, relying on the shortest-path, is the most dangerous, resulting in $194$ collisions. The ablation shows that the Refusal-Only approach performs almost identically, with $187$ collisions. This proves that simply refusing tasks does not improve safety. The safety gain comes entirely from the Causal Routing heuristic: by itself, it reduces collisions to just $80$. The Full Causal framework is marginally better at $78$ collisions, confirming that the smart heuristic---which proactively routes the robot away from busy areas---is the primary driver of safety.

This observation is confirmed by the proxemic compliance in Fig.~\ref{fig:res-safety-proxemics}. The Baseline and Refusal-Only approaches are again nearly identical: their interquartile ranges (IQRs) dip into the social zone, and their lower whiskers show frequent, close interactions in the personal and intimate zones. In contrast, both the Causal Routing and Full Causal approaches are far safer. Their median distances are higher, and their entire IQRs are maintained within the public zone. This demonstrates that the causal heuristic~(Eq.~\ref{eq:causal-inference-heuristic}) is highly effective in ensuring the robot operates at safer distances.

As done for the efficiency, we verified the statistical significance of our results, in this case using a Negative Binomial test for the human-robot collision graph and a Mann-Whitney U test for the proxemic compliance metric.

To better demonstrate the tangible benefits of our causal approach, a supplementary video showing the experimental environment created by PeopleFlow and comparing the baseline and full causal approaches in two scenarios is available online\footnote{ \url{https://drive.google.com/file/d/1oBtTpPZbstD2a66BGxG7XdfBZ84MLPM7/view?usp=sharing}}.

\section{Conclusion}\label{sec:conclusion}
In this work, we presented a novel end-to-end, causality-based decision-making framework that, for the first time, combines causal discovery from sensor data with causal reasoning for motion planning in ROS. We also introduced PeopleFlow, a new HRSI simulator featuring a mobile robot, multiple pedestrians, and diverse contextual factors that influence human and robot goals, their interactions, and the robot’s task execution.
Our approach unifies data processing, causal discovery, and causal inference into a single pipeline, enabling the robot to make better path planning decisions and to determine in advance whether to proceed with or cancel a task. These decisions are based on two causally-inferred quantities: expected people densities in specific areas of the environment, and estimated battery consumption to reach a target location.
We evaluated the proposed system in a simulated warehouse scenario, created using PeopleFlow. The scenario, inspired by our funding project DARKO, incorporates both workers and a robot with context-dependent goals, where the spatial behaviour of each is mutually influenced by the presence of the other. The results validate the improvements achieved by our solution in terms of planning efficiency and safety for shared human-robot environments.

Building on these results, 
our primary future work will focus on real-world validation. This involves deploying the framework on a real robot for onboard online decision-making, a crucial step for evaluating its effectiveness in logistics scenarios involving multiple humans and complex interactions.
We will also explore the key advantage of our end-to-end architecture: its potential for long-term autonomy. This includes enabling the robot to autonomously re-run the learning pipeline on new data to adapt to environmental changes, such as new worker schedules, without human intervention.
Another important direction is multi-robot scalability. While our current model is single-robot, the framework is designed to scale by allowing a central fleet manager to use our causal model as a component in a multi-agent heuristic for conflict-resolution. Finally, a promising direction is to enhance the learning pipeline by integrating a module for causal feature learning (i.e., which variables are important for predicting a certain outcome), which would enable the robot to automatically identify and refine relevant features, further enhancing its adaptability in dynamic environments

\section*{Acknowledgements}
This work has received funding from the European Union’s Horizon 2020 research and innovation programme under grant agreement No 101017274 (DARKO).
GB is also supported by PNRR MUR project PE0000013-FAIR.

\bibliographystyle{unsrt}
\bibliography{references}

\appendix
\section{Causal Query Scalability Analysis}\label{app:scalability}
\begin{figure}[h!]
\centering
\includegraphics[trim={0cm 0cm 0cm 0cm}, clip, width=\textwidth]{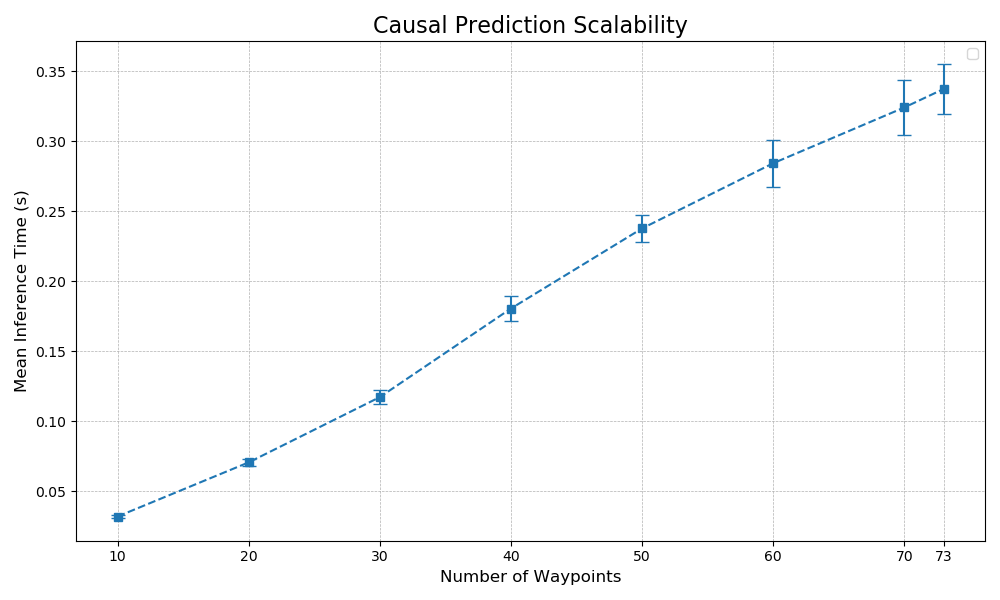}
\caption{Causal query scalability analysis.}
\label{fig:scalability}
\end{figure}
A critical requirement for any online planning system is a predictable and efficient runtime. A complex query for choosing a path, even if accurate, becomes impractical if it introduces multi-second delays or has a non-linear (e.g., exponential) cost that makes it unusable on larger maps. Therefore, we conducted a detailed scalability analysis to rigorously evaluate the computational cost of our causal query and verify its suitability for real-time applications.

Beyond the hardware configuration (specified in Sec.~\ref{sec:exp}), it is important to understand how the causal query scales with the number of waypoints. Therefore, we incrementally tested the system by generating sub-graphs of increasing size, starting from $10$ waypoints and increasing in steps of $10$ up to the full map ($73$ waypoints). For each sub-graph size, we executed the full prediction query $1000$ times, each with a randomised time slot and robot velocity.

The result of this scalability analysis is shown in Fig.~\ref{fig:scalability}, where we report the mean and standard deviation of the inference time for each sub-graph size. This analysis revealed a highly predictable linear relationship between the number of waypoints and the mean inference time. The error bars, representing the standard deviation, confirm the consistency of the query time. This predictable, linear scaling behaviour confirms that the computational cost does not explode with map complexity, which allows us to confidently predict performance even on larger maps and validates the approach for real-time planning.

\section{Heuristic Weight Sensitivity Analysis}\label{app:sensitivity}
To evaluating the impact of heuristic weight changes on decision outcomes, we performed a comprehensive sensitivity analysis. The objective was to systematically validate our parameter selection and confirm that it represents an optimal trade-off between task efficiency and robot safety.

\subsection{Methodology}
To conduct this analysis, we evaluated $5$ representative delivery tasks. These tasks were executed twice, during two distinct time slots (H2, a working time, and H6, a lunch time), to assess the framework's performance in two different scenarios of human activity.

We selected three distinct levels for each of the three heuristic weights: $\lambda_\delta$ ($0.1$, $1$, $10$), $\lambda_{D}$ ($1$, $10$, $100$), and $\lambda_{L}$ ($0.5$, $5$, $50$). This resulted in $27$ unique weight configurations, allowing us to systematically study their interactions and impact on performance. This selection of logarithmically-spaced values was chosen to test a wide range of magnitudes to thoroughly validate our intended prioritisation strategy of people ($\lambda_{D}$) over battery cost ($\lambda_{L}$), and finally over goal distance ($\lambda_\delta$).

\subsection{Results and Discussion}
The results of this analysis are summarised in Table~\ref{tab:sensitivity_analysis}. As hypothesised, the analysis confirms that the configurations yielding the best performance, under the combined efficiency and safety metrics defined in Sec. 5.3, were indeed those that adhered to this intended prioritisation.

From Table~\ref{tab:sensitivity_analysis}, we first filtered these $27$ configurations to ensure robust and safe performance. Any configuration that resulted in one or more collisions (Collis. $> 0$) or failed to achieve a $100\%$ mission success rate (Succ. $< 100\%$) was excluded from consideration (marked in red). 
From the remaining set of successful and safe configurations, we observed that configurations maintaining this $\lambda_{D} > \lambda_{L} > \lambda_\delta$ order consistently produced the strongest results. Although all configurations within this successful group yielded broadly similar, high-quality performance, we selected the parameters ($\lambda_\delta=1$, $\lambda_{D}=10$, $\lambda_{L}=5$) (marked in green) as the optimal set. This configuration, ($1$, $10$, $5$), offers the best trade-off between operational performance (i.e., task time, travelled distance, and battery usage) and safety, maintaining a significantly larger minimum distance of $0.53$~m. Overall, this parameter combination represents the optimal balance between task efficiency and human safety.

\begin{sidewaystable}[htbp]
\centering
\caption{Comprehensive sensitivity analysis of $\lambda_\delta$, $\lambda_{D}$, and $\lambda_{L}$ Parameters}
\label{tab:sensitivity_analysis}
\setlength{\tabcolsep}{4pt}
\scriptsize
\begin{tabular}{@{}lccccccccccccccccc@{}}
\toprule
\textbf{Configuration} & \textbf{Succ.} & \textbf{Collis.} & \multicolumn{4}{c}{\textbf{Time}} & \multicolumn{4}{c}{\textbf{Distance}} & \multicolumn{3}{c}{\textbf{Battery}} & \multicolumn{4}{c}{\textbf{Proxemics}} \\
\cmidrule(lr){4-7} \cmidrule(lr){8-11} \cmidrule(lr){12-14} \cmidrule(lr){15-18}
($\lambda_\delta, \lambda_{D}, \lambda_{L}$) & & & Tot. & Active & Stalled & Wasted & Tot. & Planned & Extra & Wasted & Tot. & Effective & Wasted & Median & Q1 & Q3 & Min \\
& (\%) & (count) & (h) & (\%) & (\%) & (\%) & (m) & (\%) & (\%) & (\%) & (cycles) & (\%) & (\%) & (m) & (m) & (m) & (m) \\ \midrule
(0.1, 1, 0.5) & 100.0 & 0 & 0.198 & 96.2 & 3.8 & 0.0 & 0.269 & 93.6 & 6.4 & 0.0 & 0.045 & 100.0 & 0.0 & 5.80 & 4.70 & 6.76 & 0.39 \\
\rowcolor{red!20} (0.1, 1, 5) & 70.0 & 0 & 0.198 & 63.5 & 4.0 & 32.4 & 0.231 & 67.1 & 4.9 & 28.0 & 0.044 & 67.2 & 32.8 & 5.04 & 3.62 & 6.17 & 0.32 \\
\rowcolor{red!20} (0.1, 1, 50) & 40.0 & 5 & 0.272 & 46.0 & 8.9 & 45.1 & 0.371 & 23.5 & 10.4 & 66.1 & 0.057 & 50.3 & 49.7 & 4.48 & 2.97 & 5.89 & 0.07 \\
(0.1, 10, 0.5) & 100.0 & 0 & 0.197 & 96.4 & 3.6 & 0.0 & 0.264 & 95.4 & 4.6 & 0.0 & 0.045 & 100.0 & 0.0 & 5.78 & 4.72 & 6.76 & 0.37 \\
(0.1, 10, 5) & 100.0 & 0 & 0.209 & 95.0 & 5.0 & 0.0 & 0.278 & 90.3 & 9.7 & 0.0 & 0.047 & 100.0 & 0.0 & 5.79 & 4.71 & 6.76 & 0.37 \\
\rowcolor{red!20} (0.1, 10, 50) & 70.0 & 0 & 0.199 & 64.7 & 4.0 & 31.4 & 0.234 & 66.2 & 8.6 & 25.3 & 0.044 & 68.4 & 31.6 & 4.81 & 3.44 & 6.09 & 0.37 \\
(0.1, 100, 0.5) & 100.0 & 0 & 0.203 & 97.0 & 3.0 & 0.0 & 0.280 & 96.5 & 3.5 & 0.0 & 0.047 & 100.0 & 0.0 & 5.55 & 4.49 & 6.58 & 0.31 \\
(0.1, 100, 5) & 100.0 & 0 & 0.204 & 96.2 & 3.8 & 0.0 & 0.276 & 96.0 & 4.0 & 0.0 & 0.047 & 100.0 & 0.0 & 5.63 & 4.61 & 6.64 & 0.58 \\
\rowcolor{red!20} (0.1, 100, 50) & 100.0 & 1 & 0.217 & 95.8 & 4.2 & 0.0 & 0.297 & 89.1 & 10.9 & 0.0 & 0.050 & 100.0 & 0.0 & 5.67 & 4.69 & 6.67 & 0.22 \\
\midrule
\rowcolor{red!20} (1, 1, 0.5) & 100.0 & 1 & 0.215 & 94.6 & 5.4 & 0.0 & 0.289 & 87.0 & 13.0 & 0.0 & 0.048 & 100.0 & 0.0 & 5.72 & 4.60 & 6.68 & 0.26 \\
\rowcolor{red!20} (1, 1, 5) & 70.0 & 3 & 0.183 & 67.3 & 3.2 & 29.5 & 0.223 & 69.5 & 4.4 & 26.1 & 0.041 & 70.2 & 29.8 & 5.07 & 3.62 & 6.23 & 0.09 \\
\rowcolor{red!20} (1, 1, 50) & 50.0 & 2 & 0.243 & 51.0 & 9.0 & 40.0 & 0.295 & 37.6 & 8.8 & 53.5 & 0.050 & 56.5 & 43.5 & 4.47 & 3.01 & 5.97 & 0.20 \\
(1, 10, 0.5) & 100.0 & 0 & 0.191 & 96.8 & 3.2 & 0.0 & 0.261 & 96.2 & 3.8 & 0.0 & 0.044 & 100.0 & 0.0 & 5.78 & 4.73 & 6.77 & 0.47 \\
\rowcolor{green!30}(1, 10, 5) & 100.0 & 0 & 0.193 & 96.8 & 3.2 & 0.0 & 0.262 & 96.0 & 4.0 & 0.0 & 0.044 & 100.0 & 0.0 & 5.75 & 4.71 & 6.78 & 0.53 \\
\rowcolor{red!20} (1, 10, 50) & 80.0 & 1 & 0.201 & 75.8 & 4.7 & 19.4 & 0.251 & 70.1 & 12.1 & 17.8 & 0.045 & 80.0 & 20.0 & 4.98 & 3.65 & 6.14 & 0.27 \\
\rowcolor{red!20} (1, 100, 0.5) & 100.0 & 1 & 0.206 & 95.2 & 4.8 & 0.0 & 0.275 & 91.5 & 8.5 & 0.0 & 0.047 & 100.0 & 0.0 & 5.76 & 4.69 & 6.78 & 0.08 \\
(1, 100, 5) & 100.0 & 0 & 0.192 & 97.2 & 2.8 & 0.0 & 0.267 & 94.2 & 5.8 & 0.0 & 0.045 & 100.0 & 0.0 & 5.83 & 4.75 & 6.78 & 0.52 \\
(1, 100, 50) & 100.0 & 0 & 0.199 & 95.7 & 4.3 & 0.0 & 0.267 & 94.4 & 5.6 & 0.0 & 0.045 & 100.0 & 0.0 & 5.85 & 4.75 & 6.86 & 0.42 \\
\midrule
\rowcolor{red!20} (10, 1, 0.5) & 40.0 & 0 & 0.229 & 44.1 & 8.7 & 47.1 & 0.198 & 40.4 & 15.8 & 43.8 & 0.047 & 49.0 & 51.0 & 4.38 & 2.93 & 5.88 & 0.35 \\
\rowcolor{red!20} (10, 1, 5) & 40.0 & 2 & 0.203 & 41.3 & 2.7 & 56.1 & 0.214 & 39.0 & 9.0 & 52.0 & 0.045 & 43.2 & 56.8 & 4.36 & 2.93 & 5.91 & 0.12 \\
\rowcolor{red!20} (10, 1, 50) & 10.0 & 8 & 0.188 & 8.8 & 0.3 & 90.9 & 2.864 & 0.6 & 0.1 & 99.3 & 0.041 & 9.3 & 90.7 & 4.56 & 3.02 & 5.91 & 0.01 \\
(10, 10, 0.5) & 100.0 & 0 & 0.202 & 95.1 & 4.9 & 0.0 & 0.263 & 94.4 & 5.6 & 0.0 & 0.046 & 100.0 & 0.0 & 5.92 & 4.85 & 6.85 & 0.32 \\
(10, 10, 5) & 100.0 & 0 & 0.189 & 97.4 & 2.6 & 0.0 & 0.262 & 95.9 & 4.1 & 0.0 & 0.044 & 100.0 & 0.0 & 5.76 & 4.65 & 6.76 & 0.45 \\
\rowcolor{red!20} (10, 10, 50) & 60.0 & 0 & 0.196 & 52.7 & 1.0 & 46.3 & 0.222 & 60.1 & 4.6 & 35.3 & 0.045 & 54.8 & 45.2 & 4.87 & 3.38 & 6.06 & 0.46 \\
(10, 100, 0.5) & 100.0 & 0 & 0.199 & 96.6 & 3.4 & 0.0 & 0.271 & 91.9 & 8.1 & 0.0 & 0.046 & 100.0 & 0.0 & 5.84 & 4.86 & 6.77 & 0.29 \\
\rowcolor{red!20} (10, 100, 5) & 90.0 & 1 & 0.186 & 88.1 & 3.0 & 8.9 & 0.253 & 87.6 & 3.5 & 8.8 & 0.043 & 91.1 & 8.9 & 5.77 & 4.74 & 6.78 & 0.10 \\
(10, 100, 50) & 100.0 & 0 & 0.192 & 96.6 & 3.4 & 0.0 & 0.262 & 96.1 & 3.9 & 0.0 & 0.044 & 100.0 & 0.0 & 5.73 & 4.70 & 6.82 & 0.51 \\
\bottomrule
\end{tabular}
\end{sidewaystable}

\end{document}